\documentclass[twoside]{article}

\usepackage[accepted]{aistats2024}
\usepackage{times}
\usepackage{soul}
\usepackage{url}
\usepackage[hidelinks]{hyperref}
\usepackage[utf8]{inputenc}
\usepackage[small]{caption}
\usepackage{graphicx}
\usepackage{amsmath}
\usepackage{amsthm}
\usepackage{booktabs}
\usepackage{algorithm}
\usepackage[switch]{lineno}
\usepackage{multirow}
\usepackage{enumitem}
\usepackage{nccmath}
\usepackage{amsthm}
\theoremstyle{plain}
\newtheorem{proposition}{Proposition}

\theoremstyle{definition}
\newtheorem{definition}{Definition}

\newtheorem{example}{Example}
\theoremstyle{remark}

\usepackage{cite}
\usepackage[font=footnotesize]{subfig}
\usepackage{algorithm}
\usepackage{graphicx}
\usepackage{xcolor} 
\usepackage{amssymb}
\usepackage{algpseudocode}
\usepackage[normalem]{ulem}
\usepackage{cases}

\newcommand\blfootnote[1]{%
  \begingroup
  \renewcommand\thefootnote{}\footnote{#1}%
  \addtocounter{footnote}{-1}%
  \endgroup
}

\usepackage[round]{natbib}

\usepackage{graphicx}

\begin{document}

%

%

\twocolumn[

\aistatstitle{Pathwise Explanation of ReLU Neural Networks}

\aistatsauthor{ Seongwoo Lim\textsuperscript{\rm 1} \And Won Jo\textsuperscript{\rm 2} \And  Joohyung Lee\textsuperscript{\rm 3,4} \And Jaesik Choi\textsuperscript{\rm 2,5}}



\aistatsaddress{ 
\\
\textsuperscript{\rm 1}Ulsan National Institute of Science and Technology (UNIST) \\ 
\textsuperscript{\rm 2}Korea Advanced Institute of Science and Technology (KAIST) \\
\textsuperscript{\rm 3}Arizona State University, \textsuperscript{\rm 4}Samsung Research, \textsuperscript{\rm 5}INEEJI } ]

\begin{abstract}
    Neural networks have demonstrated a wide range of successes, but their ``black box" nature raises concerns about transparency and reliability. Previous research on ReLU networks has sought to unwrap these networks into linear models based on activation states of all hidden units. In this paper, we introduce a novel approach that considers subsets of the hidden units involved in the decision making path. This pathwise explanation provides a clearer and more consistent understanding of the relationship between the input and the decision-making process. Our method also offers flexibility in adjusting the range of explanations within the input, i.e., from an overall attribution input to particular components within the input. Furthermore, it allows for the decomposition of explanations for a given input for more detailed explanations. Experiments demonstrate that our method outperforms others both quantitatively and qualitatively. 
\end{abstract}

\section{Introduction}


Neural networks have demonstrated a wide range of successes in various domains~\citep{caruana2015intelligible,litjens2017survey,yurtsever2020survey,zhu2016traffic}. However, many of these networks are perceived as ``black boxes'' due to the opacity of their decision-making processes. This has led to the rise of eXplainable Artificial Intelligence (XAI), which seeks to clarify how these black box models operate~\citep{das2020opportunities,arrieta2020explainable,adadi2018peeking}. XAI is vital in ensuring the reliability of practical applications~\citep{gunning2019xai,gunning2019darpa}, diagnosing malfunctions in neural networks~\citep{lapuschkin2019unmasking}, and promoting consistent operation in our daily lives.

\begin{figure}[!t]
    \centering
    \includegraphics[width=.85\columnwidth]{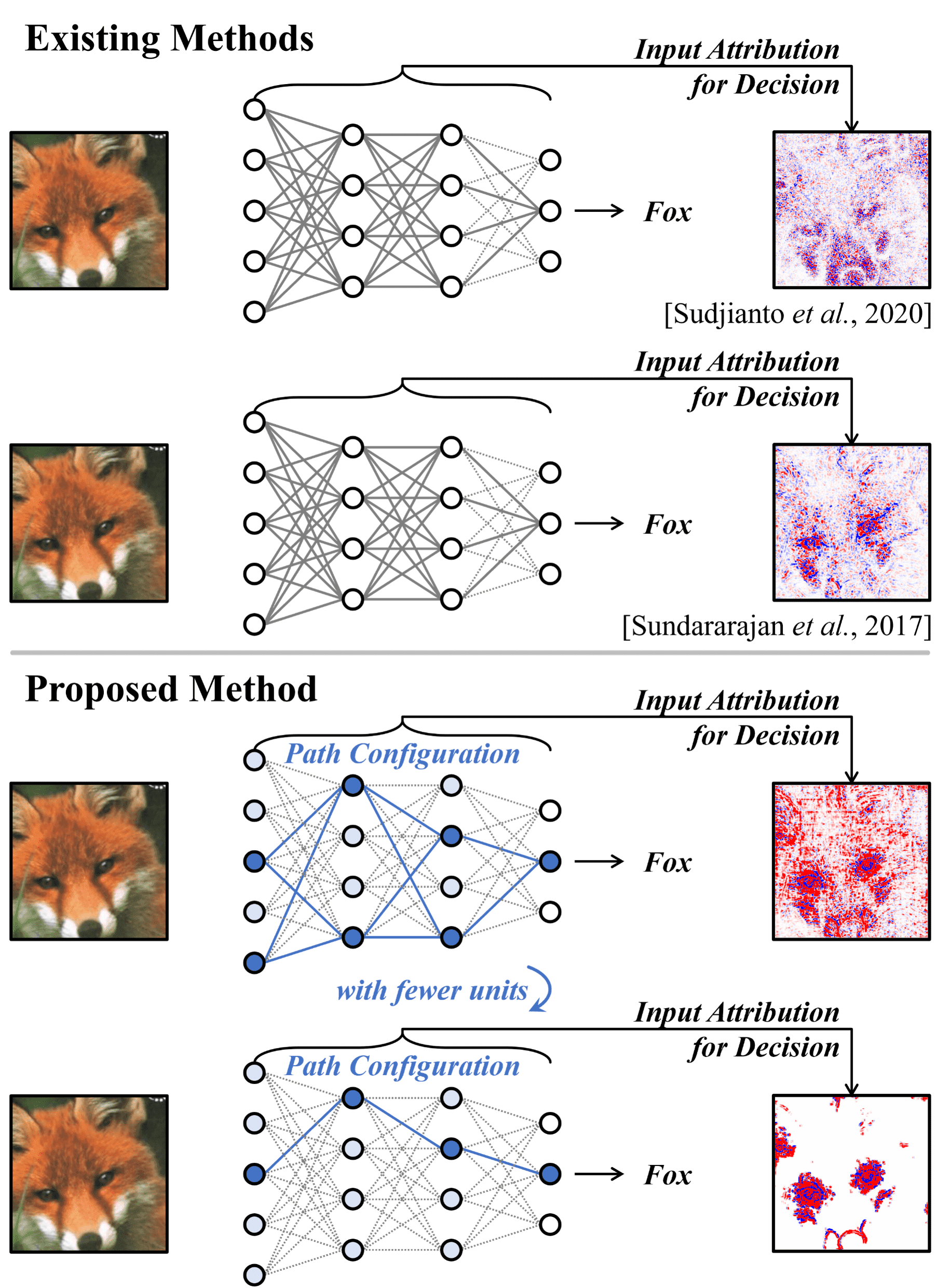}
    \vspace{-2mm}
    \caption{Comparison of the methods \vspace{-3mm}}
    \label{fig:overview}
\end{figure}

ReLU is widely employed as an activation function due to its advantages, including mitigating  the vanishing gradient problem and enabling efficient computation \citep{goodfellow2016deep}.
It is known that a Feed-Forward Neural Network with ReLU can be represented as a piecewise linear model \citep{sattelberg2020locally}.
\cite{sudjianto2020unwrapping} introduce a process called {\em unwrapping}, 
which leverages this property to describe a ReLU NN as a collection of local linear functions based on {\em activation patterns}---that is, the activation states of all hidden units within the network. Consequently, a ReLU NN is represented by a distinct linear model for inputs that correspond to the same activation pattern.\blfootnote{Code: https://github.com/Jo-won/PathwiseExplanation}

In this paper, we also represent ReLU NN as a piecewise linear model. However, instead of focusing on activation patterns, we consider activation states for individual \textit{paths}---specific subsets of hidden units involved in a particular decision process. This approach facilitates a more explicit explanation of the relationship between the input and the corresponding \textit{paths} they traverse for a decision, in contrast to prior works.\footnote{See Section~\ref{ssec:comparision} for more details.}

Additionally, the concept of a \textit{path} offers flexibility in adjusting the range of explanations within the input. Specifically, by controlling the maximum number of units included in the path configuration process for each layer, as shown in Figure~\ref{fig:overview}, our method is capable of not only providing an overall attribution of the input to the prediction but also facilitating the explanation of particular components within the input. We reveal that this adjustment can provide more detailed explanations for the decision by enabling the decomposition of explanations for a given input.

Furthermore, we demonstrate that our pathwise explanation offers better consistency in relation to the input compared to existing methodologies. This enhanced consistency arises from the utilization of only a subset of hidden units rather than the entire set. In our experiments on the use of linear models for recognizing informative attributions, our linear model derived from various types of paths outperforms others both quantitatively and qualitatively.

Our contributions are as follows:
\begin{itemize} [noitemsep,topsep=0pt,parsep=0pt,partopsep=0pt]
    \item We introduce a {\em pathwise} explanation for ReLU NNs along with its associated algorithms for computing paths. The pathwise explanation describes a clearer relationship between the input and the corresponding decision-making by the use of \textit{paths} (Section~\ref{sec:pathwise}).
    \item The pathwise explanation facilitates the decomposition of explanations for a given input (e.g. \textit{Fox}) into its individual components (e.g., \textit{Fox}'s eyes and ears). Furthermore, it elucidates the underlying reasons for incorrect predictions made by the model (Section~\ref{sec:explanation-decomposition} and~\ref{sec:explanation-for-incorrect}).
    \item In experiments recognizing informative attributions for explaining specific inputs, we demonstrate that our method with various paths outperforms others both quantitatively and qualitatively (Section~\ref{sec:quantitative-experiments}).
\end{itemize}

\section{Related Work} \label{sec:related}

As mentioned, our work is related to \cite{sudjianto2020unwrapping}. However, in contrast to their approach that considers all activated hidden units, we focus on specific subsets that form paths leading to the decision process. \citep{villani2023unwrapping} extends \textit{unwrapping} from \cite{sudjianto2020unwrapping} to NNs with diverse structures, such as Graph Neural networks and tensor convolutional networks.

Input attribution methods, designed to  compute an input's contribution to the output---either through a heatmap or as a linear model \citep{ribeiro2016should}---are widely utilized for interpreting NNs. Among them, gradient-based input attribution techniques offer intuitive means to elucidate the roles of hidden units. 

The Saliency method employs the gradient of the input for a specific target unit \citep{simonyan2013deep}. This gradient inherently signifies the input's contributions to the target unit. Moreover, it becomes possible to modify the input to enhance target units using this gradient, as described in \citep{Mordvintsev2015InceptionismGD}. The goal is often to identify input sections that amplify the target unit's response, as opposed to suppressing it. 
Techniques such as Guided Backpropagation \citep{springenberg2014striving} and Deconvnet \citep{zeiler2014visualizing} leverage the ReLU activation function to counteract the effects of negative weights. The Integrated Gradient method \citep{sundararajan2017axiomatic}, adhering to both sensitivity and implementation invariance standards, computes the integral of the gradient from a baseline to the input. Class Activation Mapping (CAM)~\citep{zhou2016learning} increases the model's transparency by using Global Average Pooling (GAP) rather than fully-connected. GradCAM~\citep{selvaraju2017grad} is an extension of CAM that can generate explanations for differentiable models without using GAP. Another class of input attribution methods revolves around perturbation-based approaches. For instance, Occlusion method \citep{zeiler2014visualizing} modifies a specific rectangular input region to a baseline and observes consequent output changes.

Some studies produce model explanations leveraging the unique properties of the ReLU function. Notably, in a neural network's Taylor series expansion with ReLU activation, explanations can be derived by solely considering the first-order term; this is because higher-order terms become zero with ReLU \citep{bach2015pixel,montavon2017explaining}.

\section{Pathwise Explanation of ReLU NN} \label{sec:pathwise}

In this section, we introduce the concept of a \textit{path} within a NN with ReLU activation.
First, we establish the idea of a \textit{one-way complete path}, representing a straightforward linear model case. Subsequently, we expand the definition to accomodate multi-way paths. 

We consider a Feed-Forward NN with a ReLU activation function in each layer, which is termed as a ReLU NN.
For an input $X\in \mathbb{R}^{d_0}$, where $d_0$ represents the vectorized input dimension, a ReLU NN $f: \mathbb{R}^{d_0} \xrightarrow[]{}  \mathbb{R}^{d_{N+1}}$ computes the prediction $f(X)$ with the ouput dimension $d_{N+1}$. For the $i^{th}$ layer, $layer_i$, a vector of hidden units $h^{(i)}=[h^{(i)}_{1},...,h^{(i)}_{d_i}]$ is given before ReLU activation, where $d_i$ is the number of hidden units in $layer_i$. A weight matrix between $layer_{i{-}1}$ and $layer_i$ is denoted as $W^{(i)}\in \mathbb{R}^{d_{i}\times d_{i{-}1}}$, and a weight associating the $j^{th}$ hidden unit in $layer_{(i{-}1)}$ and the $k^{th}$ hidden unit in $layer_{i}$ is denoted as $W^{(i)}_{k,j}\in \mathbb{R}$. The term $*$ is used as an index for all hidden units.  For example, $W^{(i)}_{*,j}\in \mathbb{R}^{d_i\times 1}$ represents weights connecting the $j^{th}$ hidden unit in $layer_{i{-}1}$ to all hidden units in $layer_i$. A bias vector for $layer_i$ is denoted as $b^{(i)}\in \mathbb{R}^{d_i}$.

A ReLU NN can be represented as an undirected graph, with hidden units serving as nodes and weights as edges. In this context, we say that a set of hidden units is {\em connected} in a ReLU NN if the subgraph they induce is connected. 

\begin{figure}[t]
\centering
\subfloat[\footnotesize{Original network}]{
   \makebox[3.8cm][c]{\includegraphics[width=0.3\columnwidth]{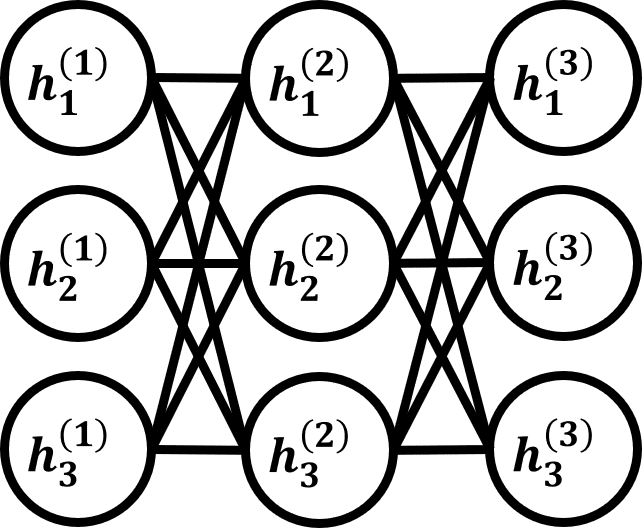}}}
   \hspace{1em}
\subfloat[One-way complete path]{
   \makebox[3.8cm][c]{\includegraphics[width=0.3\columnwidth]{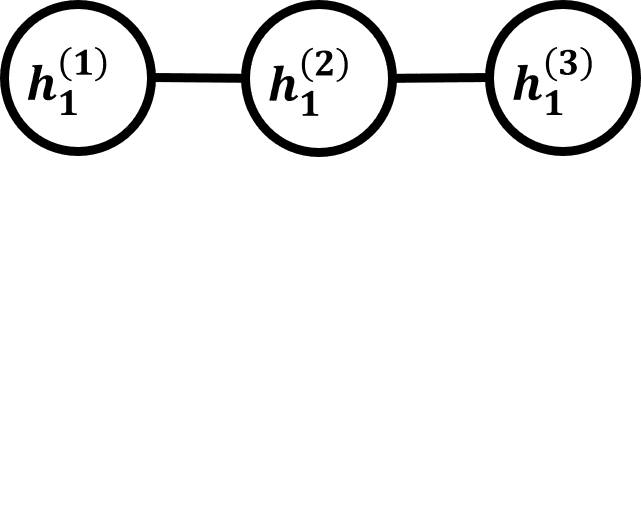}}}
\vskip\baselineskip
\subfloat[Multi-way incomplete path]{
    \makebox[3.8cm][c]{\includegraphics[width=0.3\columnwidth]{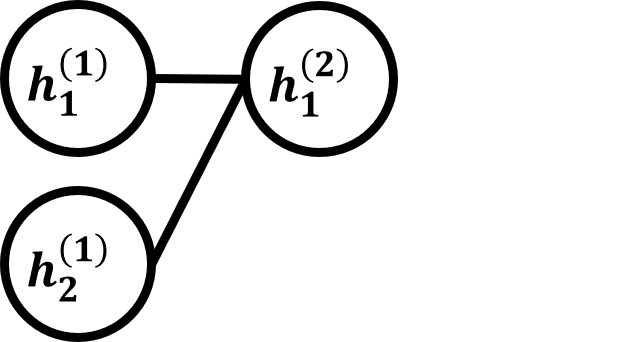}}}
\hspace{1em}
\subfloat[Not a path]{
    \makebox[3.8cm][c]{\includegraphics[width=0.3\columnwidth]{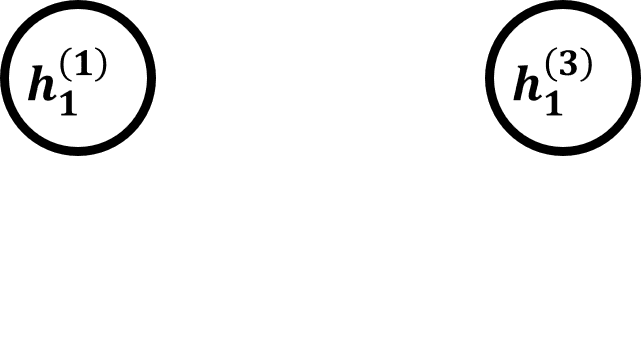}}}
    \vspace{-2mm}
\caption{Examples of paths with three hidden layers\vspace{-4mm}}
\label{fig:path example}
\end{figure}

\begin{definition} \label{def:path}
    A \textbf{\textit{path}} is defined as a set of hidden units \textit{connected} in a ReLU NN.
    \begin{itemize}[noitemsep,topsep=0pt,parsep=0pt,partopsep=0pt]
    \item A \textbf{\textit{one-way path}} is defined as a set of hidden units with no more than one unit per layer (in contrast to a \textbf{\textit{multi-way path}}).
    \item A \textbf{\textit{complete path}} is defined as a set of hidden units with at least one unit in every layer (in contrast to an \textbf{\textit{incomplete path}}).
    \item A \textbf{\textit{path is activated}} when all units in the path have positive values with regard to the input; otherwise, the \textbf{\textit{path is deactivated}}.
    \item The \textbf{\textit{depth of a path}} is the number of layers that include hidden units in the path.
    \end{itemize}
\end{definition}

\begin{example}
    For the ReLU NN with 3 hidden layers in Figure~\ref{fig:path example} (a), $\{h^{(1)}_{1},h^{(2)}_{1},h^{(3)}_{1}\}$ represents a \textit{one-way complete path} with a depth of 3, 
    and $\{h^{(1)}_{1},h^{(1)}_{2},h^{(2)}_{1}\}$ is a \textit{multi-way incomplete path} with a depth of 2. Note that $\{h^{(1)}_{1},h^{(3)}_{1}\}$ is not a \textit{path} because $h^{(1)}_{1}$ and $h^{(3)}_{1}$ are not connected. 
\end{example}

\subsection{Formalization of \textit{One-way Complete Path}} \label{sec:one-way-complete-path}

The ReLU function can be expressed as \useshortskip
\begin{linenomath}
    \begin{align*}
        ReLU(x) = x \cdot \phi(x),
    \end{align*}
    \end{linenomath}
\useshortskip
where $\phi(x)=1$ for $x > 0$, $\phi(x)=0$ for $x \leq 0$.  $\phi$ can be applied elementwise for a multi-dimensional input. A Feed-Forward ReLU NN can be represented as follows: \useshortskip
\begin{linenomath}
    \begin{align}
        h^{(1)} &= W^{(1)}X +b^{(1)} \nonumber\\
        h^{(2)} &= W^{(2)}(h^{(1)}\cdot \phi(h^{(1)})) +b^{(2)} \nonumber\\
        \dots \label{eq:our ReLU full equation}\\
        h^{(N)} &= W^{(N)}(h^{(N-1)}\cdot \phi(h^{(N-1)})) + b^{(N)} \nonumber\\
        f(X) &= W^{(N+1)}(h^{(N)}\cdot \phi(h^{(N)})) +b^{(N+1)} .\nonumber
    \end{align}
\end{linenomath}
For each hidden unit, Equation~\eqref{eq:our ReLU full equation} can be decomposed:
\begin{linenomath}
    \begin{align}
        h^{(1)}_{i_1} &= W^{(1)}_{i_1,*}X +b^{(1)}_{i_1}\nonumber \\
        h^{(n)}_{i_n} &= \sum^{d_{n{-}1}}_{k=1} W^{(n)}_{i_n,k} h^{(n{-}1)}_{k}\phi (h^{(n{-}1)}_{k}) +b^{(n)}_{i_n}\nonumber \\
        f(X) &= \sum^{d_{N}}_{k=1} W^{(N+1)}_{*,k}h^{(N)}_{k}\phi (h^{(N)}_{k}) +b^{(N{+}1)},  \nonumber
    \end{align}
\end{linenomath}
where $0{<}i_1 {\in} \mathbb{N}{\leq} d_{1}$, $0{<}i_n {\in} \mathbb{N}{\leq} d_{n}$ and $1{<}n{\in} \mathbb{N}{<}N{-}1$. Furthermore, $f(X)$ can be {\em unfolded} as \useshortskip
\begin{linenomath}
\begin{align}
    \begin{split}
    &f(X) = \\
    &  \sum_{i_1,i_2,..,i_N}W^{(N+1)}_{*,i_N}W^{(N)}_{i_N,i_{N{-}1}}\dots W^{(2)}_{i_2,i_1}W^{(1)}_{i_1,*}X\prod_{j=1}^N\phi(h^{(j)}_{i_j})\\
    &+ \sum_{i_1,i_2,..,i_N}W^{(N+1)}_{*,i_N}W^{(N)}_{i_N,i_{N{-}1}}\dots W^{(2)}_{i_2,i_1}b^{(1)}_{i_1}\prod_{j=1}^N\phi(h^{(j)}_{i_j})\\
    &+ \sum_{i_2,..,i_N}W^{(N+1)}_{*,i_N}W^{(N)}_{i_N,i_{N{-}1}}\dots W^{(3)}_{i_3,i_2}b^{(2)}_{i_2}\prod_{j=2}^N\phi(h^{(j)}_{i_j})\\
    &\dots \\
    &+ \sum_{i_N}W^{(N+1)}_{*, i_N}b^{(N)}_{i_N}\prod_{j=N}^N\phi(h^{(j)}_{i_j}) + b^{(N{+}1)}, 
    \end{split} \label{eq:ReLU-NN-equation}
\end{align}
\end{linenomath}
\useshortskip
where $0{<}i_n {\in} \mathbb{N}{\leq} d_{n}$.

One of the primary concepts in this paper is representing a ReLU NN as a piecewise linear model based on a path derived from Equation~\eqref{eq:ReLU-NN-equation}. The ReLU function, due to its piecewise linear nature, allows for the expression that is a sum of piecewise linear models comprised of $\phi$ for each hidden unit. Given a one-way complete path ${\bf p} = [h^{(1)}_{i_1},...,h^{(N)}_{i_N}]$, we define the terms $W^{\bf p}$ and $b^{\bf p}$ as follows: \useshortskip
\begin{linenomath}
    \begin{align}
            W^{\bf p} &= W^{(N+1)}_{*,i_{N}}W^{(N)}_{i_{N},i_{(N-1)}}\dots W^{(2)}_{i_{2},i_{1}}W^{(1)}_{i_{1},*}\nonumber \\
            b^{\bf p} &= W^{(N+1)}_{*,i_{N}}W^{(N)}_{i_{N},i_{(N-1)}}\dots W^{(2)}_{i_{2},i_{1}}b^{(1)}_{i_{1}}, \label{eq:bpath} 
    \end{align}
\end{linenomath}
where  $1{\le} i_j{\in} \mathbb{N} {\le} d_j$. Note that in Equation~\eqref{eq:ReLU-NN-equation}, these $W^{{\bf p}}$ and $b^{{\bf p}}$ are multiplied by $\prod_{h\in {\bf p}}\phi(h)$---the product of $\phi$ for all hidden units in ${\bf p}$. As a result, we define the {\em piecewise linear model $f^{\bf p}(X)$ for the path~${\bf p}$}~as \useshortskip
\begin{linenomath}
    \begin{align}
        f^{\bf p}(X)=\left(W^{\bf p}X + b^{\bf p}\right)\prod_{h\in {\bf p}}\phi(h). \label{ eq:plm}
    \end{align}
\end{linenomath}
\useshortskip
\begin{proposition} \label{prop:one-way-bpath}
    For a one-way complete path ${\bf p}$, the piecewise linear model $f^{\bf p}(X)$ constitutes a term in Equation~\eqref{eq:ReLU-NN-equation} and is non-zero if and only if the path is activated (the proof is in Appendix~\ref{sec:proof-one-way-bpath}). 
\end{proposition}

\subsection{Generalization of Linear Model for Multi-way Complete Path}

The preceding section discussed a one-way complete path. Given the piecewise linear nature of a ReLU NN, the extension to multi-way complete paths is straightforward. As previously, we define the piecewise linear model for a multi-way complete path ${\bf p}$ in the following form:\useshortskip
\begin{linenomath}
    \begin{align}
        f^{\bf p}(X) = \left(W^{\bf p}X+b^{\bf p}\right)\prod_{h\in {\bf p}}\phi(h) \label{eq:form of linear model}
    \end{align}
\end{linenomath}
where $W^{\bf p} \in \mathbb{R}^{d_{N+1} \times d_0}$, $b^{\bf p} \in \mathbb{R}^{d_{N+1}}$. 

For a multi-way complete path ${\bf p}$ for a ReLU NN, we define 
$W^{\bf p}$ and $b^{\bf p}$ for equation (4) by summing up the weights and biases of all possible one-way complete paths that are subsets of the multi-way complete path~${\bf p}$.\useshortskip

\begin{linenomath}
    \begin{align}
        W^{\bf p} = \sum_{\substack{\text{$p$ is a one-way complete path} \label{eq:multi-way-weight} \\ \text{that is a subset of {\bf p}}}} W^{p}  \\ 
        b^{\bf p} = \sum_{\substack{\text{$p$ is a one-way complete path} \label{eq:multi-way-bias} \\ \text{that is a subset of {\bf p}}}} b^{p} 
    \end{align}
\end{linenomath}

\begin{proposition}\label{prop:multi-way-complete-path}
For a multi-way complete path ${\bf p}$, the piecewise linear model $f^{\bf p}(X)$ represents a summation of terms from Equation~\eqref{eq:ReLU-NN-equation}, and is non-zero if and only if the path is activated. 
\end{proposition}

\begin{figure}[t]
     \centering
     \includegraphics[width=0.75\columnwidth]{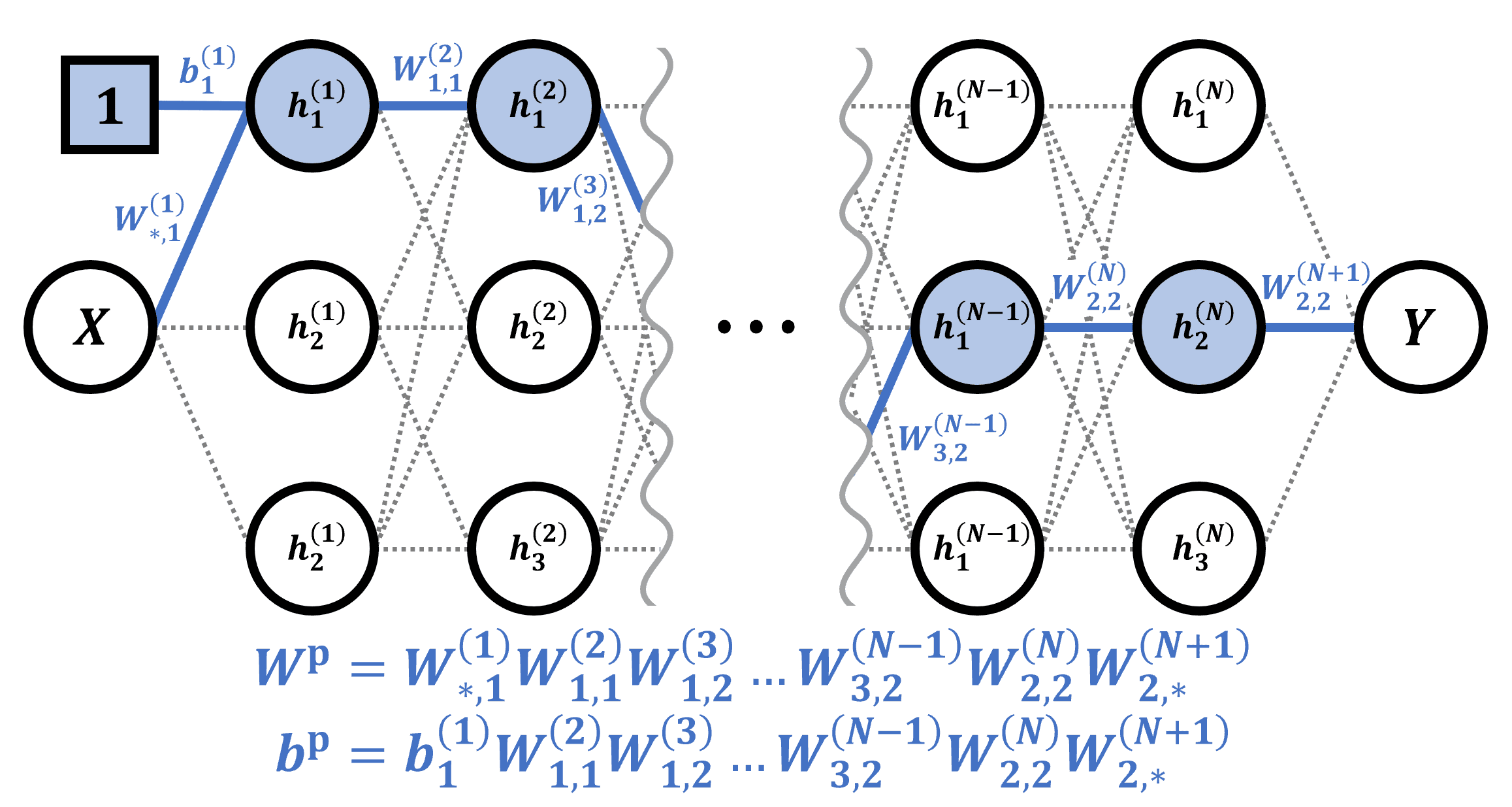}
     \vspace{-2mm}
    \caption{An illustration of Proposition~\ref{prop:one-way-bpath}} 
    \label{fig: one-way complete path illustration}
\end{figure}
\begin{figure}[t]
     \centering
     \includegraphics[width=0.9\columnwidth]{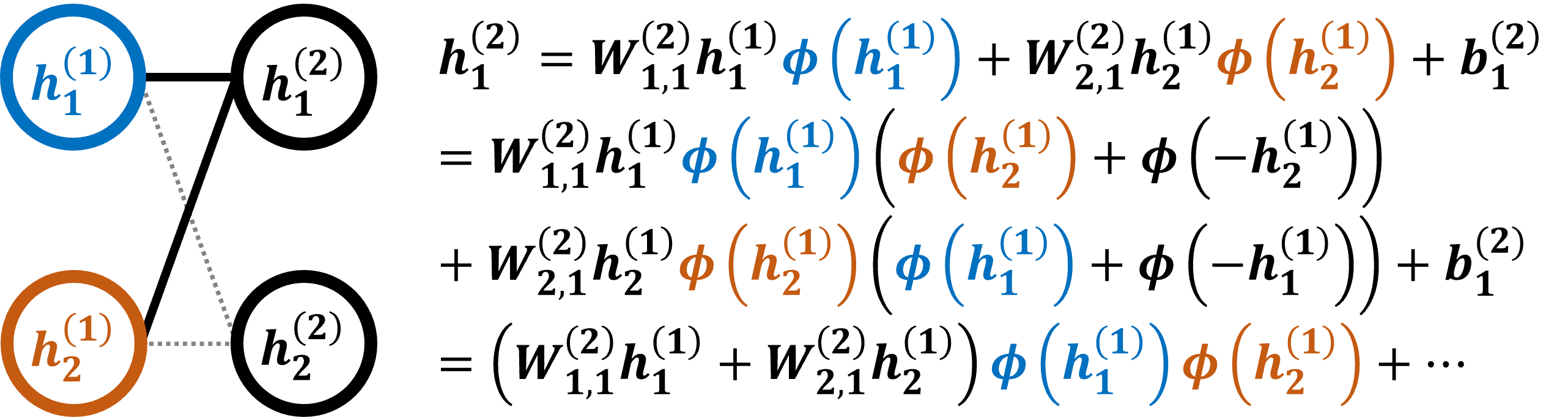}
     \vspace{-2mm}
    \caption{{An illustration of Proposition~\ref{prop:multi-way-complete-path}}}
    \label{fig:multi-way-complete-path-illustration}
\end{figure}

The proof of Proposition~\ref{prop:multi-way-complete-path} (Appendix~\ref{sec:proof-multiway-complete-path}) uses the following property of a ReLU function: \useshortskip
\begin{linenomath}
    \begin{align}
        \phi(x) + \phi(-x) = 1 \label{eq:relu-two-states-property}
    \end{align}
\end{linenomath}
which means that the hidden unit can have two cases: activated ($\phi(x){=}1$) or deactivated ($\phi(-x){=}1$). Figure~\ref{fig:multi-way-complete-path-illustration} illustrates Proposition~\ref{prop:multi-way-complete-path}.

\begin{example}
As in Figure~\ref{fig:multi-way-complete-path-illustration}, consider the multi-way complete path ${\bf p}=[h^{(1)}_1, h^{(1)}_2, h^{(2)}_1, h^{(2)}_2]$. \useshortskip
\begin{linenomath}
    \begin{align*}
    W^{\bf p} 
         =&  \left(\frac{df(x)}{dh^{(2)}_{1}}\frac{dh^{(2)}_1}{dh^{(1)}_1} + \frac{df(x)}{dh^{(2)}_{2}}\frac{dh^{(2)}_2}{dh^{(1)}_1}\right)\frac{dh^{(1)}_1}{dx} \\
        &+ \left(\frac{df(x)}{dh^{(2)}_{1}}\frac{dh^{(2)}_1}{dh^{(1)}_2} + \frac{df(x)}{dh^{(2)}_{2}}\frac{dh^{(2)}_2}{dh^{(1)}_2}\right)\frac{dh^{(1)}_2}{dx}\\
        = & W^{[h^{(1)}_1\!, h^{(2)}_1]}\!+\!W^{[h^{(1)}_1\!, h^{(2)}_2]}\!+\!W^{[h^{(1)}_2\!, h^{(2)}_1]}\!+\!W^{[h^{(1)}_2\!, h^{(2)}_2]}.
    \end{align*} 
\end{linenomath}
\end{example}

\begin{figure*}[t!]
\centering
\subfloat[ReLU NN structure ]{
         \makebox[4cm][c]{\includegraphics[width=0.25\textwidth]{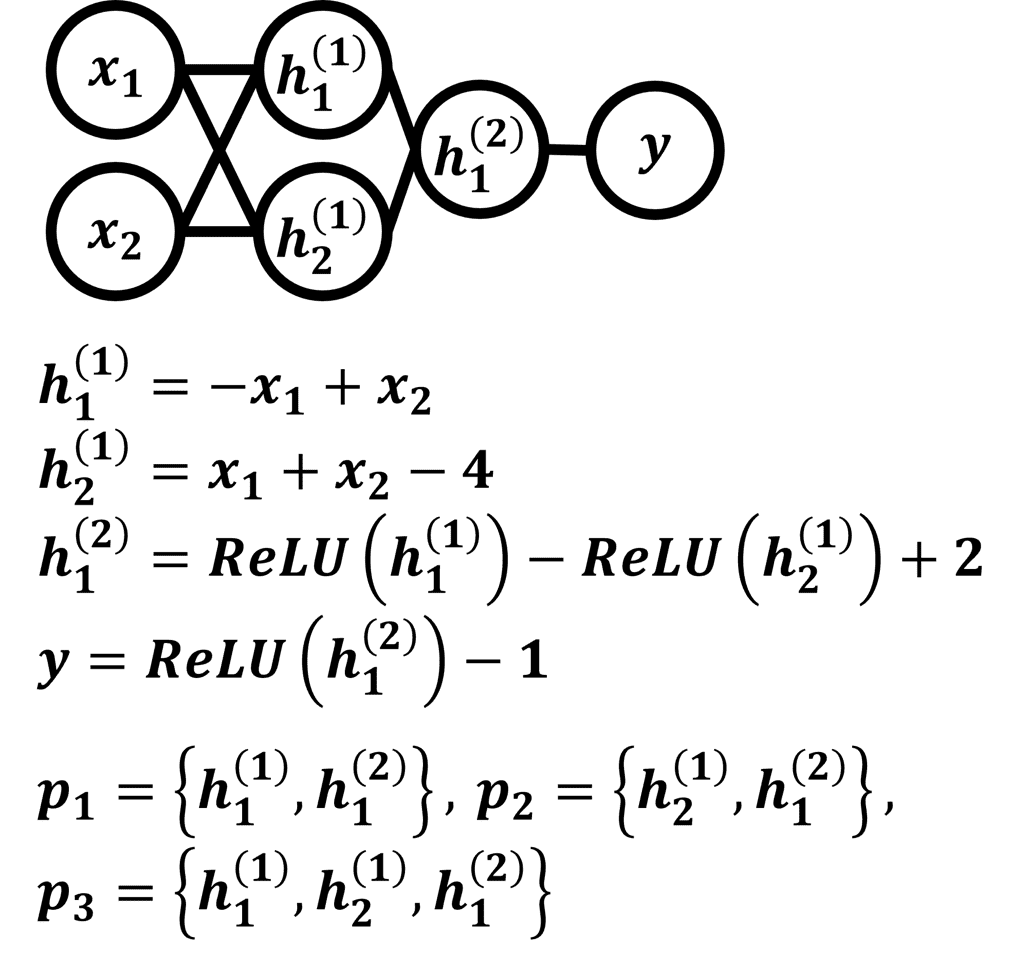}}
        }
        \hfill
\subfloat[Inconsistent explanation for the white sample as \textit{positive} class ($y{>}0$) by using $f^{{\bf p}_3}{=}f^{{\bf p}_1}{+}f^{{\bf p}_2}$]{
         \makebox[6cm][c]{\includegraphics[width=0.3\textwidth]{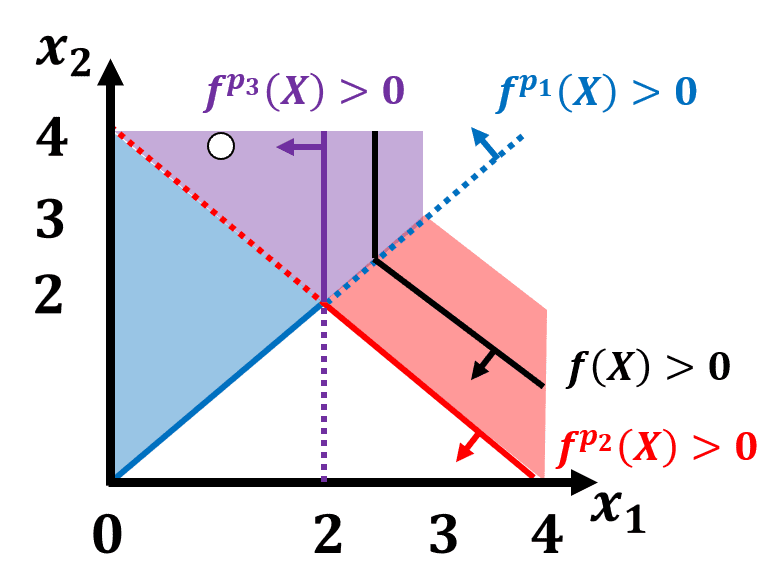}}
        }
        \hfill
\subfloat[Consistent explanation for the white sample as \textit{positive} class by using $f^{{\bf p}_1}$ only.]{
         \makebox[6cm][c]{\includegraphics[width=0.33\textwidth]{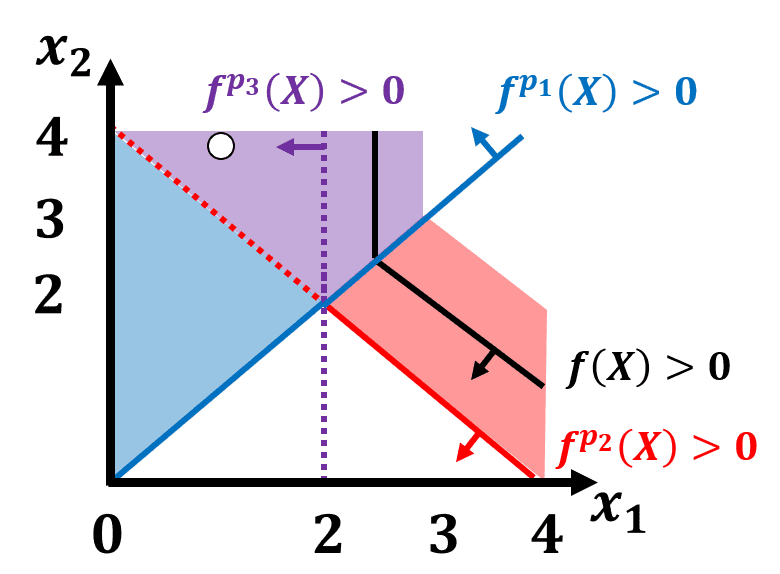}}
        }
    \vspace{-2mm}
    \caption{Comparison of the proposed pathwise explanation and \textit{unwrapping} through a pedagogical example. For the white sample, $f^{{\bf p}_1}$ and $f^{{\bf p}_2}$ predict different classes. A pathwise explanation can generate a more consistent explanation. In (b) and (c), a solid line represents a decision boundary relevant to that example, while a dotted line is included for reference purposes but does not directly relate to the example.
    }
    \label{fig:toy-example-region-vesus-our}
\end{figure*}

\begin{algorithm}[t!]
   \caption{Multi-way Path Construction}
   \label{alg:Path construction}
\begin{algorithmic}[1]
    \State {\bfseries Input:} $h$: set of hidden units in $N$ layer ReLU NN, \\
    \qquad $h^{(N+1)}$: NN output, $targetIndex$: target index, \\ \qquad $depth$: depth of the generated path\\
    \qquad $width$: width of the generated path\\
    \qquad $\alpha$: threshold for importance
    \State {\bfseries Initialize:} $path^{(N+1)}=\left[1,2,\dots,|h^{(N+1)}|\right]$\\ \quad \qquad \quad \quad $W^{prev}{=}I$ \Comment{identity matrix}
    \For{$n \gets N$ to \textcolor{black}{$(N{-}depth{+}1)$}}
        \State $path^{(n)}=emptyList()$
        \State $W=\frac{d}{dh^{(n)}}h^{(n+1)}$ \Comment{$\mathbb{R}^{d_{n+1}{\times} d_n}$}
        \State $W = W^{prev}(W[path^{(n+1)},*])$ \Comment{$\mathbb{R}^{d_{N+1}{\times} d_n}$}
        \State $imp=matMul(W, diag(h^{(n)}))^T$ \Comment{$\mathbb{R}^{d_n {\times} d_{N+1}}$}
        \State $imp = softmax(imp)[*, targetIndex]$ \Comment{$\mathbb{R}^{d_{n}}$}
        \State $setOfTopKIndex {=} topKIndex(imp, \textcolor{black}{K{=}width})$
        \For{$index$ in $setOfTopKIndex$}
            \If{$imp[index] > \alpha$}
                \State $path^{(n)}.add(h^{(n)}_{index})$
            \EndIf
        \EndFor
        \State $W^{prev} = W[*, path^{(n)}]$ \Comment{$\mathbb{R}^{d_{N+1}\times |path^{(n)}|}$}
    \EndFor
\State \Return{$\cup_{n=(N{-}depth{+}1)}^{N}path^{(n)}$}
\end{algorithmic}
\end{algorithm}
\begin{example} Consider the ReLU NN, as in Figure~\ref{fig:toy-example-region-vesus-our}. There exist three complete paths, ${\bf p}_1{=}\{h^{(1)}_1, h^{(2)}_1\}$, ${\bf p}_2{=}\{h^{(1)}_2, h^{(2)}_1\}$ and ${\bf p}_3{=}\{h^{(1)}_1, h^{(1)}_2, h^{(2)}_1\}$. Then, \useshortskip
\begin{linenomath}
    \begin{align*}
        f^{{\bf p}_1}(x_1,x_2)&=(-x_1+x_2)\phi(h^{(1)}_1)\phi(h^{(2)}_1), \\
        f^{{\bf p}_2}(x_1,x_2)&=(-x_1-x_2+4)\phi(h^{(1)}_2)\phi(h^{(2)}_1), \\
        f^{{\bf p}_3}(x_1,x_2)&=(-2x_1+4)\phi(h^{(1)}_1)\phi(h^{(1)}_2)\phi(h^{(2)}_1).
    \end{align*}
\end{linenomath}
The ReLU NN can be expressed as\useshortskip
\begin{linenomath}
    \begin{align*}
        f(x_1,x_2)&=f^{{\bf p}_1}(x_1,x_2)+f^{{\bf p}_2}(x_1,x_2)+2\phi(h^{(2)}_1)-1\\
        &=
        \begin{cases}
            f^{{\bf p}_1}(x_1,x_2)+1, & \mbox{in the  blue region} \\
            f^{{\bf p}_2}(x_1,x_2)+1, & \mbox{in the red region} \\
            f^{{\bf p}_3}(x_1,x_2)+1, & \mbox{in the purple region}. 
        \end{cases}
    \end{align*}
\end{linenomath}
The explanation by $f^{{\bf p}_1}(x_1,x_2)$ means that $x_1$ has a negative contribution towards the classification of the \textit{positive} class ($y{>}0$), whereas $x_2$ has a positive contribution. Note that $f^{{\bf p}_3}(x_1,x_2){=}f^{{\bf p}_1}(x_1,x_2)+f^{{\bf p}_2}(x_1,x_2)$ within the purple region. Our proposed method enables explanations from three distinct paths: $p_1$, $p_2$, and $p_3$, individually for inputs within this purple region. 
\end{example}

\subsection{Path construction}\label{sec:path-construction}

This section presents an algorithm to find paths that are positively related to the target class prediction. Algorithm~\ref{alg:Path construction} is a pseudocode for our path construction method. We compute the weight between the $(n{+}1)^{th}$ layer and the $n^{th}$ layer by gradient (line 10). The importance of the hidden units in the $n^{th}$ layer is computed by $softmax(W^{(N+1)}h^{(N)})$ (lines 12--13). The top-k important units are those that increases the target class prediction value over predictions for other classes (lines 14--19). \textcolor{black}{The $depth$, $width$, $\alpha$ are hyperparameters used to generate various paths. $width$ indicates the maximum number of hidden units used at each layer when constructing a path.} $\alpha$ controls how much the prediction value for the target class is increased by the hidden units compared to other classes. \textcolor{black}{We consistently set $\alpha$ at $\frac{1}{|classes|}$, where $|classes|$ represents the total number of classes in the dataset. This approach is adopted to identify neurons that contribute significantly more to the target class than the average contribution across all classes, thereby highlighting neurons with a substantial impact on the decision-making process.}

\textcolor{black}{
The path generated by Algorithm~\ref{alg:Path construction} consists of hidden units from the $(N{-}depth{+}1)^{th}$ layer to the $N^{th}$ layer.
Then we may consider that the ReLU NN is partitioned into two subnetworks: $f_1$ that spans from the first layer to the $(N{-}depth{+}1)^{th}$ layer, and $f_2$ that spans from the $(N{-}depth{+}1)^{th}$ layer to the $N^{th}$ layer. The path that is found by Algorithm~\ref{alg:Path construction} is incomplete for the original NN, but is complete for $f_2$.
Thus, we can compute the linear model for this path. 
For the first subnetwork, we can employ direct linearization: \useshortskip
\begin{linenomath}
    \begin{align*}
        W_1&=\frac{d}{dX}f_1(X),\\
        b_1&= f_1(X) - W_1X.
    \end{align*}
\end{linenomath}
Here, $f_1(X)$ denotes the first subnetwork, which is ReLU NN itself. We then obtain the linear model of the whole ReLU NN for the incomplete path ${\bf p}$ as a composite function of these linear models, i.e., $f_2^{{\bf p}}(W_1X+b_1)$. 
}

\begin{example} \label{ex:path-construction-algorithm}
Let's revisit the simple ReLU NN in Figure~\ref{fig:toy-example-region-vesus-our}. We will show how the paths are constructed by the algorithm~\ref{alg:Path construction} (with \textcolor{black}{$depth{=}2$ and $width{=}1$}) for the white sample $(x_1,x_2){=}(1.0, 4.0)$ in Figure~\ref{fig:toy-example-region-vesus-our} (b) as the \textit{positive} class~($y{>}0$). For the white sample, \useshortskip
\begin{linenomath}
    \begin{align*}
        h^{(1)}_1=3,\quad h^{(1)}_2=1,\quad h^{(2)}_1=4,\quad y=3
    \end{align*}
\end{linenomath}
\useshortskip
For the second layer, $W\!{=}\!\begin{bmatrix}1\end{bmatrix}$.
Then, the importance of $h^{(2)}_1$:\useshortskip
\begin{linenomath}
    \begin{align*}
        imp_{h^{(2)}_1}=\frac{e^{1\times h^{(2)}_1}}{e^{1\times h^{(2)}_1}+e^{-(1\times h^{(2)}_1)}}=0.982
    \end{align*}
\end{linenomath}
We add the $h^{(2)}_1$ to $path^{(2)}$. Next, for the first layer,\useshortskip
\begin{linenomath}
    \begin{align*}
        W &= \begin{bmatrix}
    1  \end{bmatrix}
    \begin{bmatrix}
    1 & -1  \end{bmatrix}= \begin{bmatrix}
    1 & -1  \end{bmatrix}.
    \end{align*}
\end{linenomath}
Then, we compute the importance of $h^{(1)}_1$ and $h^{(1)}_2$:\useshortskip
\begin{linenomath}
\begin{align*}
    imp_{h^{(1)}_1}=\frac{e^{1\times h^{(1)}_1}}{e^{1\times h^{(1)}_1}+e^{-(1\times h^{(1)}_1)}}=0.953, \\
    imp_{h^{(1)}_2}=\frac{e^{-1\times h^{(1)}_2}}{e^{-1\times h^{(1)}_2}+e^{-(-1\times h^{(1)}_2)}}=0.269. 
\end{align*}
\end{linenomath}
We add $h^{(1)}_1$ to $path^{(1)}$. The constructed path for the white sample is $path^{(1)}\!\cup\!path^{(2)}{=}\{h^{(1)}_1\!,\!h^{(2)}_1\}$ (Figure~\ref{fig:toy-example-region-vesus-our} (c)).
\end{example}

\section{Comparison with Unwrapping by Sudjianto \textit{et al.}} \label{ssec:comparision}

Both our pathwise explanation and \textit{unwrapping} \citep{sudjianto2020unwrapping,villani2023unwrapping} use the activation states in a ReLU NN to represent the ReLU NN as a piecewise linear model. The fundamental distinction is that the pathwise explanation leverages the activation states of a subset of hidden units connected by paths, whereas \textit{unwrapping} encompasses the activation states of all hidden units. The local linear model produced by \textit{unwrapping} for an input aligns with the linear model of the path that includes all activated hidden units for that input along with an additional bias. 
\begin{example}
Consider again the simple ReLU network in Figure~\ref{fig:toy-example-region-vesus-our} (a). According to~\cite{sudjianto2020unwrapping},  the input space is divided by activation patterns. For example, any input instance within the purple region in Figure~\ref{fig:toy-example-region-vesus-our} (b) activates $h^{(1)}_1{\And} h^{(1)}_2 {\And} h^{(2)}_1$.
Figure~\ref{fig:toy-example-region-vesus-our} (b) show a local linear model on each activation region. The weights of the linear model are used to generate explanations for the input's contribution to the prediction.
However, this method often creates inconsistent explanations.
For instance, the white input in Figure~\ref{fig:toy-example-region-vesus-our} (c) is classified as the \textit{positive} class ($y>0$) by $f^{{\bf p}_1}(x_1,x_2)+1$. However, the same input is classified as \textit{negative} class ($y<0$) by $f^{{\bf p}_2}(x_1,x_2)+1$. Consequently, the explanation by $f^{{\bf p}_1}(x_1,x_2)$ describes how the ReLU NN perceives the input as the \textit{positive} class, while the explanation by $f^{{\bf p}_2}(x_1,x_2)$ describes the opposite. Thus, the explanation generated by $f^{{\bf p}_3}(x_1,x_2)$, which is the summation of $f^{{\bf p}_1}(x_1,x_2)$ and $f^{{\bf p}_2}(x_1,x_2)$, conveys mixed messages, representing both the \textit{positive} and \textit{negative} classification. However, as explained in Example~\ref{ex:path-construction-algorithm}, our method offers a consistent explanation for predicting the white input as the \textit{positive} class by using $f^{{\bf p}_1}(x_1,x_2)$ instead of $f^{{\bf p}_3}(x_1,x_2)$ even for the purple region (Figure~\ref{fig:toy-example-region-vesus-our} (c)). 
\end{example}

The Figure~\ref{fig:overview} shows the difference in explanations generated by \textit{unwrapping} and pathwise explanation. From this figure, we can see that our method better captures the features of the input, such as the eyes and ears of the fox.

\section{Experiments} \label{sec:experiments}

This section evaluates the effectiveness of our pathwise explanation method. First, we demonstrate that multiple paths, or subsets of hidden units, can account for explanations of individual components within a given input by decomposing the model's decision-making process. In addition, it specifies the reasons for the incorrect responses generated by the model. Subsequently, we assess both the quantitative and qualitative explanations for the input, as calculated using our method.

We conducted experiments on a curated subset of the ImageNet dataset~\citep{russakovsky2015imagenet}, which includes 10 distinct classes. For the purpose of explaining decisions, regardless of the method employed, all experiments consistently utilized the VGG-16 architecture~\citep{simonyan2015very}. This architecture was fine-tuned using pretrained weights from ImageNet, with adjustments made to its classifier to match the number of classes in the subset. 

Notably, our method is applicable to CNNs, as both convolution and pooling operations are linear for a given input. Specifically, we can extend our method designed for MLPs to CNNs by equating a combined convolutional and pooling operation, $f:\mathbb{R}^{H_i\times W_i \times C_i}$, to a linear layer in an MLP, $g:\mathbb{R}^{H_i W_i C_i} \to \mathbb{R}^{H_o W_o C_o}$, which takes a flattened input and produces a flattened output. This approach has been utilized in all our experiments with CNNs, e.g., VGG-16 as detailed in Section~\ref{sec:experiments} and ResNet-18 in Section~\ref{sec:appendix_resnet}.

We compare our method with the following methods. 
\begin{itemize}[parsep=1pt] 
\item \textbf{Saliency} \citep{simonyan2013deep},
\item Input$\times$Gradient~(\textbf{IxG})~\citep{shrikumar2016not},
\item Integrated Gradients~(\textbf{IGs})~\citep{sundararajan2017axiomatic},
\item Guided Backprop.~(\textbf{GBP})~\citep{springenberg2014striving},
\item Guided GradCAM~(\textbf{GGC})~\citep{selvaraju2017grad},
\item Blur Integrated Gradients~(\textbf{BlurIG})~\citep{xu2020attribution}.
\end{itemize}
Note that the attributions provided by Saliency and IxG are exactly the same with the weight~($W$) and $WX$ of the piecewise linear model obtained by \textit{unwrapping}, respectively. The attribution map, generated by the pathwise explanation of path ${\bf p}$ for input $X$, is constructed as $W^{{\bf p}}X$.

\begin{figure}[t]
    \centering
     \includegraphics[width=0.4\textwidth]{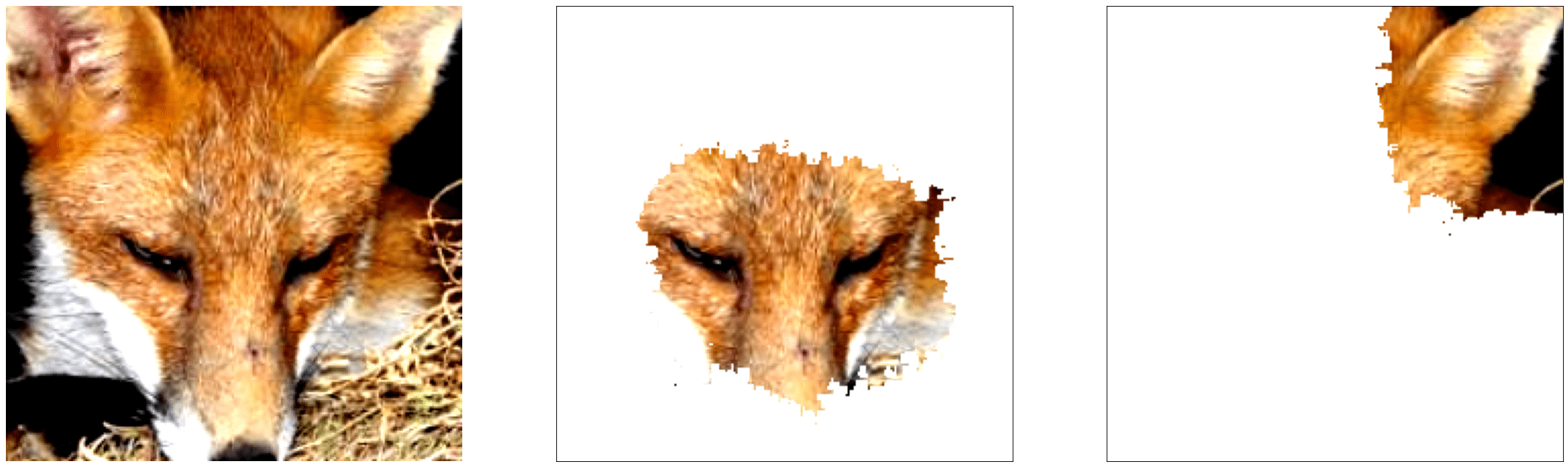}  
     \vspace{-2mm}
     \caption{An example of explanation decomposition through the paths. (left) \textit{Fox} class input; (middle) an explanation by the path using the \textit{Fox}'s eyes for the \textit{Fox} class prediction; and (right) using the \textit{Fox}'s ear. \vspace{-0mm}}
     \label{fig:visualization-multiple-features}
\end{figure}

\begin{figure}[t]
    \centering
     \includegraphics[width=0.4\textwidth]{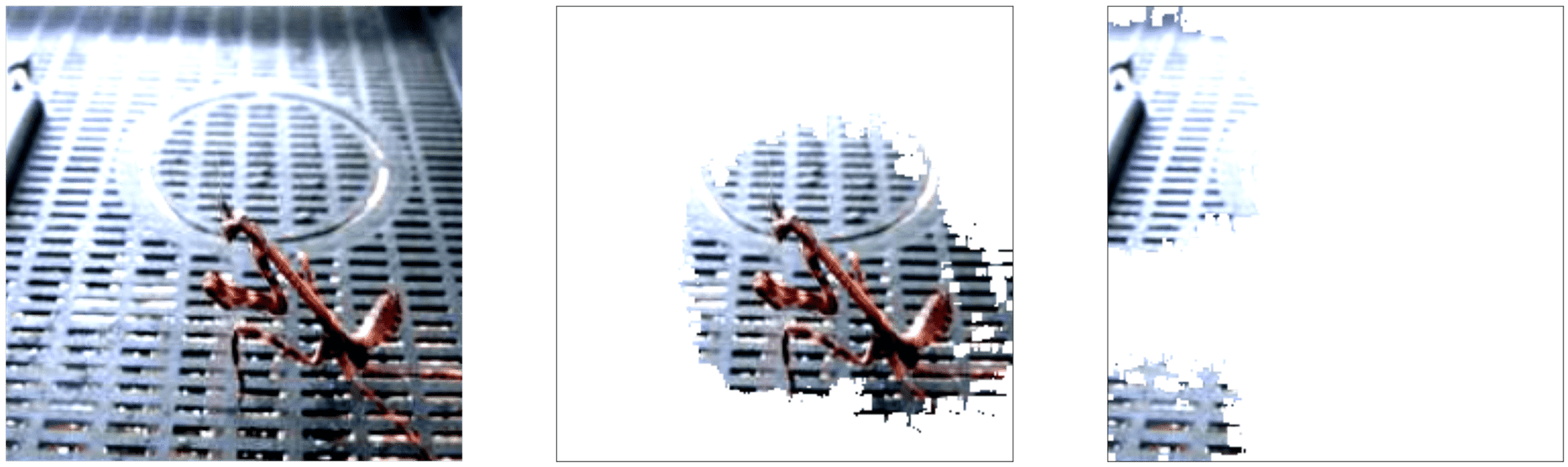}
     \vspace{-2mm}
     \caption{An example of the explanation for the incorrect prediction through the paths. (left) \textit{mantis} class input; (middle) input region supporting the ReLU prediction as \textit{mantis}; (right) input region supporting the ReLU prediction as \textit{prison}. \vspace{-0mm}}
     \label{fig:visualization-incorrect-prediction}
\end{figure}

\begin{figure*}[!t]
    \centering
    \includegraphics[width=0.8\textwidth]{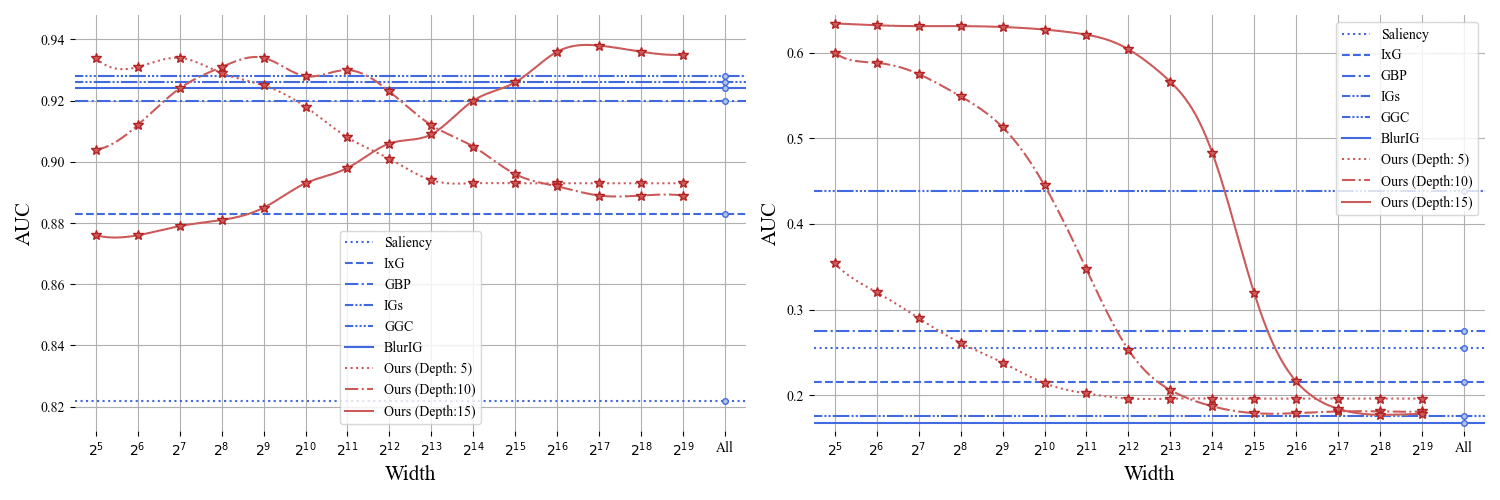}
    \vspace{-2mm}
    \caption{Quantitative explanation via \textit{insertion} and \textit{deletion} depending on the maximum number of units per layer (\textit{width}) and the number of layers (\textit{depth}) for path configuration. In our method, a \textit{depth} of 15, which is the maximum, represents a complete path, while any lesser value indicates an incomplete path.\vspace{-2mm}}
    \label{fig:ins_perf_by_wd }
\end{figure*}

\subsection{Explanation Decomposition by Paths} \label{sec:explanation-decomposition}
A significant advantage of our pathwise explanation is its ability to identify multiple distinct paths that play a crucial role in the model's decision-making. Moreover, we observed that individual paths can effectively capture salient features within the input. \textcolor{black}{Assuming that neurons in the higher layers have high-level features like the fox's eye, we generated multiple paths that include each hidden neuron extracted in Algorithm 1 from the $N^{th}$ layer. For example, if a $path^{(N)} {=} \{h_1, h_2\}$ is generated at step $n{=}N$ in Algorithm 1, we assumed $path^{(N)} {=} \{h_1\}$ and proceeded to the next step to create a path. We also assumed $path^{(N)} {=} \{h_2\}$ and proceeded to the next step to create another path.
}
In Figure~\ref{fig:visualization-multiple-features}, we present an illustrative example that showcases the model making an accurate prediction. In this instance, our proposed method identifies two distinct paths that adeptly capture essential features, such as the \textit{Fox}'s eyes and ear, vital for correctly recognizing the input as a \textit{Fox}. Moreover, our pathwise explanation facilitates the breakdown of each object within the input, allowing for the identification and explanation of multiple objects (Appendix~\ref{appx:multiple-object}).

\subsection{Explanation for Incorrect Prediction} \label{sec:explanation-for-incorrect}

Our path construction algorithm (Algorithm~\ref{alg:Path construction}) is capable of generating explanations for any class. Particularly for incorrect predictions in ReLU NNs, these explanations offer valuable insights into the underlying reasons for the model's misclassifications. This is illustrated in Figure~\ref{fig:visualization-incorrect-prediction}. In this figure, the middle image represents our explanation of the correct class (\textit{mantis}), which the model predicted with the second-highest confidence. In contrast, the right image represents our explanation for the incorrect class (\textit{prison}), which the model predicted with the highest confidence. These explanations suggest that the model's inaccurate predictions arise from similarities between the background of the input image and prison bars. Based on these observations, we conclude that the proposed explanation method can also be used to pinpoint the factors leading to the model's misclassifications.

\subsection{Pixel Insertion and Deletion Game} \label{sec:quantitative-experiments}

\begin{table}[h!]
    \centering
    \begin{tabular}{c|cc}
         & \shortstack{\textit{Insertion}($\uparrow$)} & \shortstack{\textit{Deletion}($\downarrow$)} \\
         \hline
        Saliency & 0.822 & 0.255 \\
        IxG & 0.883 & 0.215 \\
        IGs & 0.926 & 0.176 \\
        GBP & 0.920 & 0.275 \\
        GGC & 0.928 & 0.438 \\
        BlurIG & 0.924 & \textbf{0.168} \\
        \hline
        Ours & \textbf{0.936} & 0.179 \\
    \end{tabular}
    \vspace{-2mm}
    \caption{Area Under Curve (AUC) of \textit{Insertion} and \textit{Deletion}. $\uparrow$ indicates that the larger the value, the better the explanation, whereas $\downarrow$ indicates the opposite.}
    \label{tab:pixel insertion and deletion game}
\end{table}

To compare our attribution map with other explanation methods, we employed two causal metrics: \textit{insertion} and \textit{deletion}~\citep{petsiuk2018rise}. In this context, the attribute map offers a quantified explanation of the impact that each input region (or each pixel, in the case of images) has on the decision. For our method, this influence is computed using the linear model for the path. 

These two causal metrics determine that the attribution map provides a better explanation when there is a more significant degree of change according to addition or removal as described in the map. In particular, \textit{insertion} perceives a higher value as indicative of a better explanation, where the value is the Area Under Curve~(AUC) for performance changes caused by increasing the proportion of inserting important pixels. Conversely, \textit{deletion} interprets a lower value as a more proper explanation, obtained when increasing the proportion of removal.

As evidenced in Table~\ref{tab:pixel insertion and deletion game}, our method outperforms the rest in the \textit{insertion} metric and offers competitive performance in the \textit{deletion} metric. Such results underscore that our approach, distinct from others, excels at explaining all components within the input. Moreover, it facilitates elucidating each individual component, as described before.
\begin{figure*}[!t]
    \centering
    \includegraphics[width=0.75\textwidth]{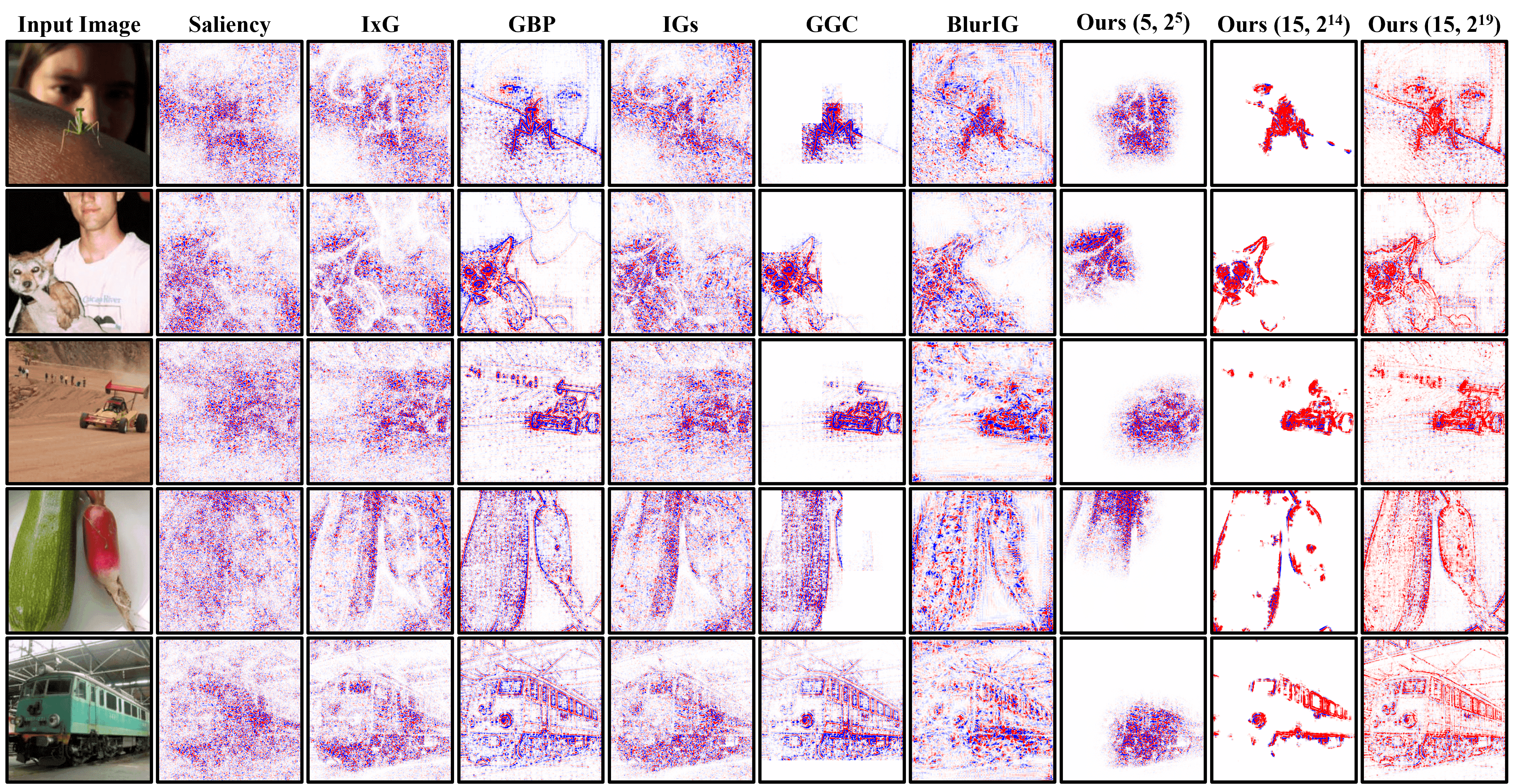}
    \vspace{-2mm}
    \caption{Qualitative explanation depending on the maximum number of units per layer (\textit{width}) and the number of layers (\textit{depth}) for path configuration. The expression enclosed in parentheses in the proposed method is as follows: (\textit{depth}, \textit{width}). Positive and negative attributions are indicated in red and blue. vspace{-2mm}}
    \label{fig:visualization special case}
\end{figure*}

\subsection{Explanation by Various Types of Path} \label{sec:various-explanation-hyperparameter}

\textcolor{black}{Various paths can be constructed by varying \textit{depth} and \textit{width} using Algorithm~\ref{alg:Path construction}.} Figures~\ref{fig:ins_perf_by_wd } and~\ref{fig:visualization special case} demonstrate that the pathwise explanations for these various paths outperform other methods both quantitatively and qualitatively. 

More precisely, in case of a small \textit{depth}, since the path includes only high-level layers containing hidden units with relatively broad receptive fields, the path with a small \textit{width} can sufficiently consider the entire area within an image. However, the quality of attribution produced by the path is degraded when the \textit{width} increases, since leads to the inclusion of less significant units. 

Conversely, in case of a large \textit{depth}, a large \textit{width} can enhance the upper bound increment of explanation by considering the entire area within an image. Figure~\ref{fig:visualization special case} graphically presents our attribution maps as influenced by both \textit{width} and \textit{depth}. Interestingly, even when both \textit{depth} and \textit{width} are minimal, the pathwise explanation aptly pinpoints the object's location. For paths with the maximum depth, our attribution maps not only stand out more than those of other methods employing all units but also adeptly highlight key features as the \textit{width} is adjusted.

\section{Conclusion}

We proposed a method of explaining a ReLU network in terms of piecewise linear model that corresponds to the \textit{path}---the subset of hidden units. We demonstrated that the proposed method can generate various explanations for a single input by employing different paths. Additionally, we introduced a path construction algorithm that generates consistent explanations for the model's output. The experiment indicates that the pathwise explanation 
provides a clearer and more consistent understanding of the relationship between the input and the decision-making process over the others.

\vfill\null

\subsubsection*{Acknowledgements}
This work was partially supported by Institute of Information \& communications Technology Planning \& Evaluation (IITP) grant funded by the Korea government(MSIT) (No.2022-0-00984, Development of Artificial Intelligence Technology for Personalized Plug-and-Play Explanation and Verification of Explanation; No.2019-0-00075, Artificial Intelligence Graduate School Program (KAIST)), the National Science Foundation under Grant IIS-2006747, and Samsung Research.


\bibliographystyle{named}

\newpage
\appendix
\onecolumn

\setcounter{figure}{0}
\setcounter{equation}{0}
\setcounter{table}{0}

\renewcommand\thefigure{\Alph{figure}}
\renewcommand\theequation{\roman{equation}}
\renewcommand\thetable{\Alph{table}}

\section{Proofs}

\subsection{Proof of Proposition~\ref{prop:one-way-bpath}} \label{sec:proof-one-way-bpath}

Given a one-way complete path ${\bf p} = [h^{(1)}_{i_1},...,h^{(N)}_{i_N}]$, we define the terms $W^{\bf p}$ and $b^{\bf p}$ as follows:\useshortskip
\begin{align}
        W^{\bf p} &= W^{(N+1)}_{*,i_{N}}W^{(N)}_{i_{N},i_{(N-1)}}\dots W^{(2)}_{i_{2},i_{1}}W^{(1)}_{i_{1},*}\nonumber \\
        b^{\bf p} &= W^{(N+1)}_{*,i_{N}}W^{(N)}_{i_{N},i_{(N-1)}}\dots W^{(2)}_{i_{2},i_{1}}b^{(1)}_{i_{1}}, \label{eq:bpath_appendix} 
\end{align}
where  $1{\le} i_j{\in} \mathbb{N} {\le} d_j$. Thus, we define the {\em piecewise linear model $f^{\bf p}(X)$ for the path ${\bf p}$ } as \useshortskip  
\begin{align}
    f^{\bf p}(X)=\left(W^{\bf p}X + b^{\bf p}\right)\prod_{h\in {\bf p}}\phi(h). \label{eq:plm}
\end{align}
\useshortskip
\noindent
{\bf Proposition~\ref{prop:one-way-bpath}.}\ \ 
{\it For a one-way complete path ${\bf p}$, the piecewise linear model $f^{\bf p}(X)$ constitutes a term in Equation~\eqref{eq:ReLU-NN-equation} and is non-zero if and only if the path is activated and $\left(W^{\bf p}X + b^{\bf p}\right)$ is non-zero. }
\begin{proof}
From Equation~\eqref{eq:ReLU-NN-equation}, we take the terms including $\prod_{h\in {\bf p}}\phi(h)$ where ${\bf p} = [h^{(1)}_{i_1},...,h^{(N)}_{i_N}]$.\useshortskip
\begin{align*}
&\left(W^{(N+1)}_{*,i_N}W^{(N)}_{i_N,i_{N{-}1}}\dots W^{(2)}_{i_2,i_1}W^{(1)}_{i_1,*}X + W^{(N+1)}_{*,i_N}W^{(N)}_{i_N,i_{N{-}1}}\dots W^{(2)}_{i_2,i_1}b^{(1)}_{i_1}\right)\prod_{j=1}^N\phi(h^{(j)}_{i_j})\\
&=\left(W^{\bf p}X + b^{\bf p}\right)\prod_{h\in {\bf p}}\phi(h).
\end{align*}
\useshortskip
By Definition~\ref{def:path}, ``the path is activated'' means that all hidden units in the path are activated given input $X$. 
Therefore,\useshortskip
\begin{align*}
    &f^{{\bf p}}(X)= \left(W^{\bf p}X + b^{\bf p}\right)\prod_{h\in {\bf p}}\phi(h)\text{ is non-zero}\\
    &\Rightarrow \prod_{h\in {\bf p}}\phi(h)=1\\
    &\Rightarrow {\bf p} \text{ is activated}.
\end{align*}
Furthermore, if the path ${\bf p}$ is activated,\useshortskip
\begin{align*}
    &\text{for any $h\in {\bf p}$, }\phi(h)=1\\
    &\Rightarrow \prod_{h\in {\bf p}}\phi(h)=1\\
    &\Rightarrow f^{{\bf p}}(X)=\left(W^{\bf p}X + b^{\bf p}\right)\prod_{h\in {\bf p}}\phi(h) \text{ is non-zero if $\left(W^{\bf p}X + b^{\bf p}\right)$ is non-zero.}
\end{align*}
\end{proof}

\subsection{Proof of Proposition~\ref{prop:multi-way-complete-path}} \label{sec:proof-multiway-complete-path}

For a multi-way complete path ${\bf p}$ for a ReLU NN, we define 
$W^{\bf p}$ and $b^{\bf p}$ by summing up the weights and biases of all possible one-way complete paths that are subsets of the multi-way complete path~${\bf p}$.\useshortskip
\begin{align}
    W^{\bf p} = \sum_{\substack{\text{$p$ is a one-way complete path} \label{eq:multi-way-weight} \\ \text{that is a subset of {\bf p}}}} W^{p}  \\ 
    b^{\bf p} = \sum_{\substack{\text{$p$ is a one-way complete path} \label{eq:multi-way-bias} \\ \text{that is a subset of {\bf p}}}} b^{p}. 
\end{align}
We define the {\em piecewise linear model $f^{\bf p}(X)$ for the path ${\bf p}$ } as \useshortskip  
\begin{align}
    f^{\bf p}(X)=\left(W^{\bf p}X + b^{\bf p}\right)\prod_{h\in {\bf p}}\phi(h). \label{ eq:plm}
\end{align}

\noindent
{\bf Proposition~\ref{prop:multi-way-complete-path}} \\ 
{\it 
For a multi-way complete path ${\bf p}$, the piecewise linear model $f^{\bf p}(X)$ represents a summation of terms from Equation~\eqref{eq:ReLU-NN-equation}, and is non-zero if and only if the path is activated and $\left(W^{\bf p}X + b^{\bf p}\right)$ is non-zero.
}

\begin{figure}[t!]
     \centering
     \includegraphics[width=0.5\columnwidth]{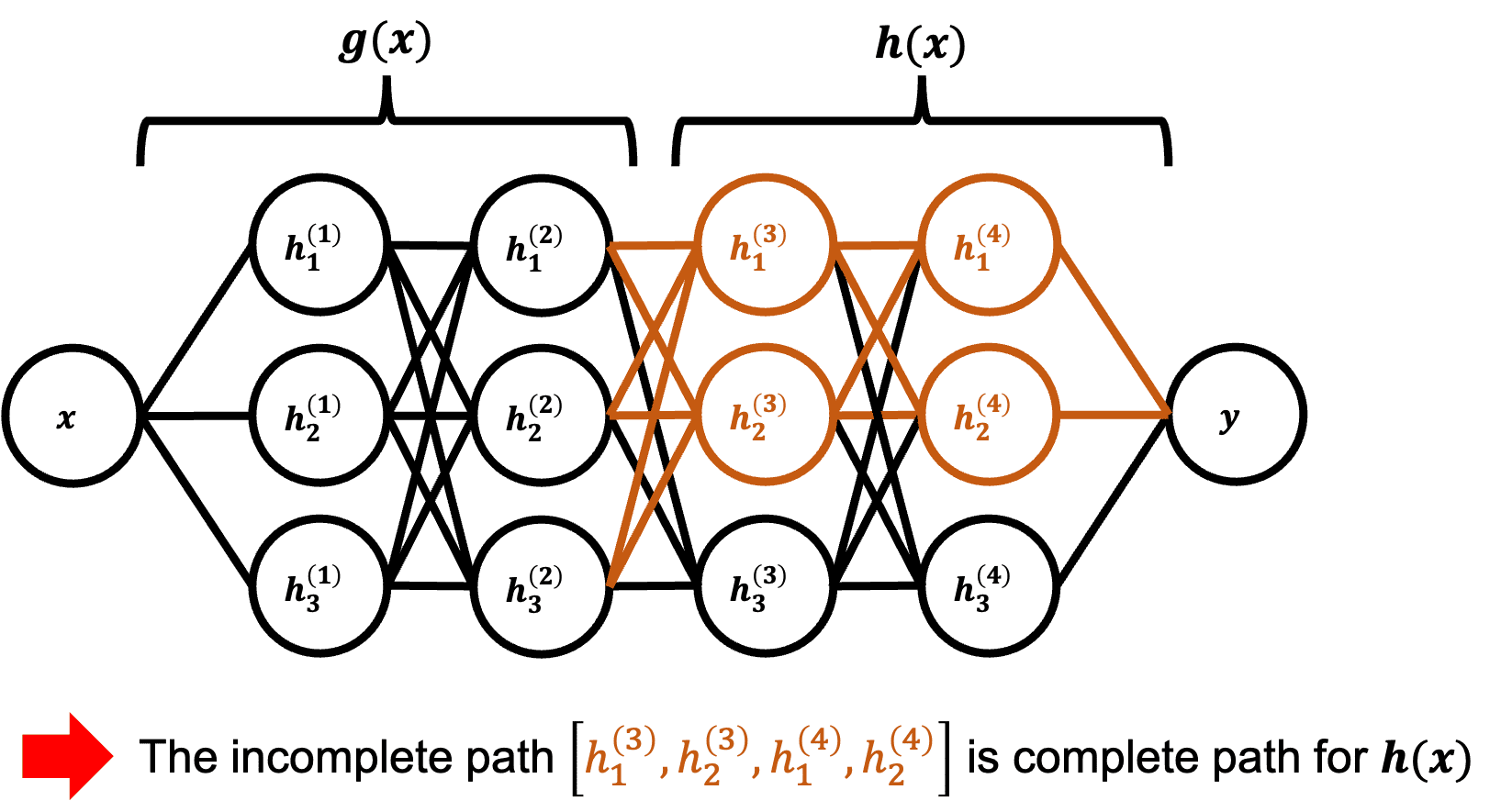}
    \caption{Example of incomplete path with \textit{depth} 2 \vspace{-2mm}}
    \label{fig:incomplete-path-illustration}
\end{figure}

\begin{proof}
    In a multi-way complete path, there exist hidden units from the same hidden layer. Without loss of generality, let $h_i$ and $h_j$ be the hidden units in the multi-way complete path from same hidden layer. From Equation~\eqref{eq:our ReLU full equation}, we can substitute $\phi(h_i)$ as $\phi(h_i)\phi(h_j) + \phi(h_i)\phi(-h_j)$ and $\phi(h_j)$ as $\phi(h_j)\phi(h_i) + \phi(h_j)\phi(-h_i)$ because $\phi(a) + \phi(-a)=1$. In other words, the decomposition from $\phi(h_i)$ to $\phi(h_i)\phi(h_j)$ and $\phi(h_i)\phi(-h_j)$ means that the $\phi(h_i)\phi(h_j)$ indicates both  $h_i$ and $h_j$ are activated, while $\phi(h_i)\phi(-h_j)$ indicates only $h_i$ is activated. In both cases, $h_i$ is activated ($\phi(h_i)$). \useshortskip
    \begin{align}
        &W_i\phi(h_i)+W_j\phi(h_j) \nonumber\\
        &= W_i\phi(h_i)\left(\phi(h_j){+}\phi(-h_j)\right){+}W_j\phi(h_j)\left(\phi(h_i){+}\phi(-h_i)\right)\nonumber \\
        &=(W_i+W_j)\phi(h_i)\phi(h_j)+\alpha(h_i, h_j), \nonumber 
    \end{align}
    where $\alpha(h_i,h_j)$ is the rest of the term except $(W_i+W_j)\phi(h_i)\phi(h_j)$.
\end{proof}

\section{Experiment Environment} \label{appx:experiment-env}
All experiments were conducted under uniform computing conditions, leveraging a single Quadro RTX 6000 GPU, running on Ubuntu 18.04, with Cuda 10.2 and Pytorch 1.11.0.

\section{Baseline Codes Used}  \label{appx:code-asset}

\begin{figure}[t]
    \centering
    \subfloat[]{
        \includegraphics[width=0.15\textwidth]{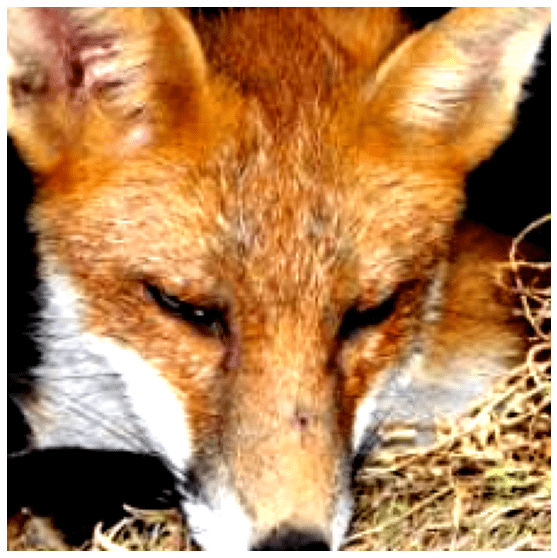}
         }
    \subfloat[]{
         \includegraphics[width=.15\textwidth]{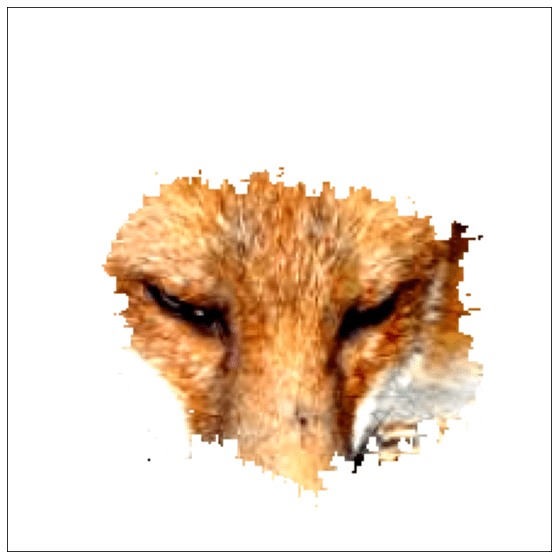}
         }
         \subfloat[]{
         \includegraphics[width=.15\textwidth]{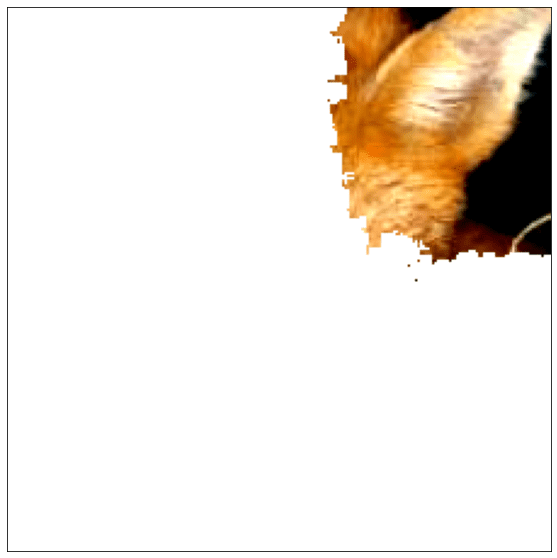}
         }
         \subfloat[]{
         \includegraphics[width=.15\textwidth]{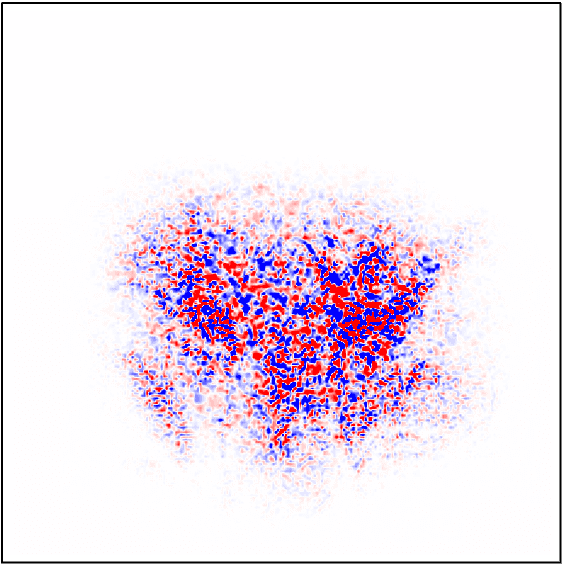}
         }
         \subfloat[]{
         \includegraphics[width=.15\textwidth]{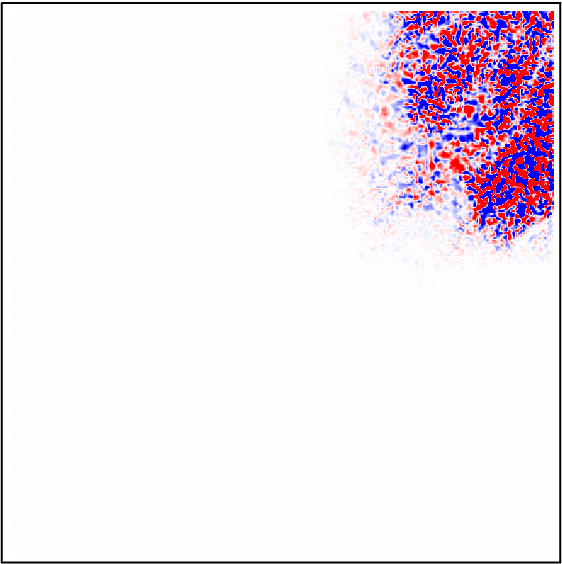}
         }
         \vspace{-2mm}
         \caption{An example of capturing multiple features through the paths. (a) \textit{fox} class input; (b) an explanation by the path using the \textit{fox}’s eyes for the \textit{fox} class prediction; and (c) using the \textit{fox}’s ear. (d, e) attribution maps for (b) and (c)\vspace{-2mm}}
         \label{fig:visualization-multiple-features-with-attribution}
\end{figure}

\begin{itemize}
    \item We used the \textbf{Captum}\footnote{https://captum.ai} package and \textbf{PAIR Saliency}\footnote{https://pair-code.github.io/saliency} package for \textbf{Saliency}, \textbf{Input x Gradient}, \textbf{Integrated Gradients}, \textbf{Guided Backpropagation}, \textbf{Guided GradCAM}, and \textbf{Blur Integrated Gradients}.
    \item We used the \textbf{RISE} \citep{petsiuk2018rise}\footnote{https://github.com/eclique/RISE/tree/master} code for the insertion and deletion game (Section~\ref{sec:quantitative-experiments} in the main script).
    \item We used the training code\footnote{https://github.com/kuangliu/pytorch-cifar} for the model, as introduced in Section~\ref{further_comp}, on CIFAR10. 
\end{itemize}

\clearpage

\section{Supplementary Material for Section~\ref{sec:explanation-decomposition}}
\label{appx:multiple-object}

\begin{figure}[t]
    \centering
    \subfloat[]{
        \includegraphics[width=0.15\textwidth]{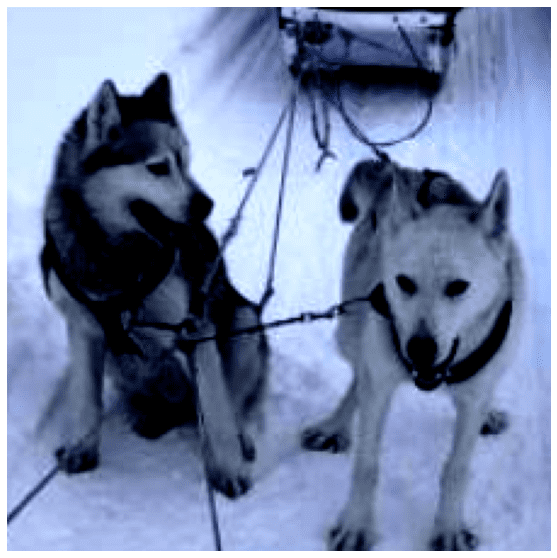}
         }
    \subfloat[]{
         \includegraphics[width=.15\textwidth]{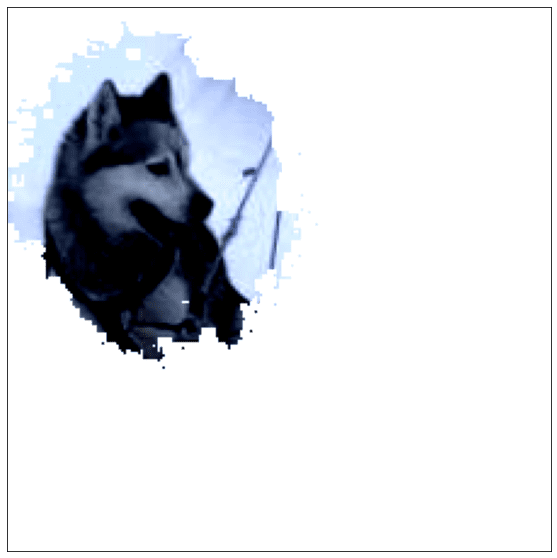}
         }
         \subfloat[]{
         \includegraphics[width=.15\textwidth]{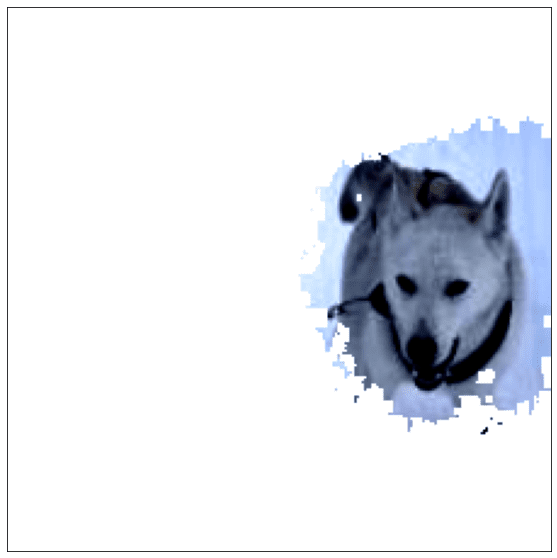}
         }
         \subfloat[]{
         \includegraphics[width=.15\textwidth]{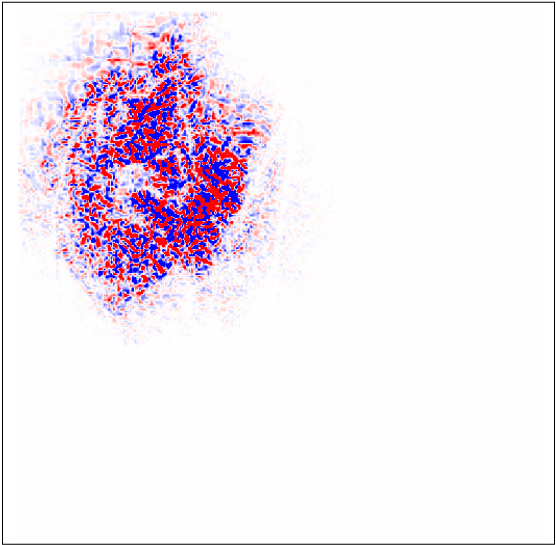}
         }
         \subfloat[]{
         \includegraphics[width=.15\textwidth]{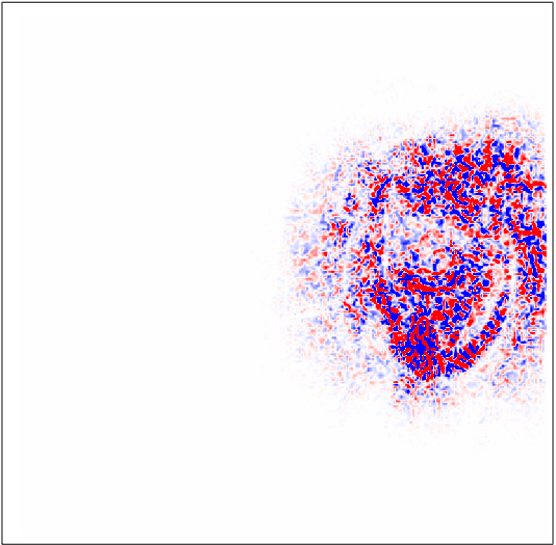}
         }
         \vspace{-2mm}
         \caption{An example of capturing multiple objects through the paths. (a) \textit{husky} class input (b) The path that explains the \textit{husky} class prediction using the left husky and (c) right husky. (d, e) attribution maps for (b) and (c)\vspace{-2mm}}
         \label{fig:visualization-multiple-objects-with-attribution}
\end{figure}

As previously discussed in Section~\ref{sec:explanation-decomposition}, our pathwise explanation is capable of generating diverse explanations for a single input by considering multiple paths. In this supplementary section, we aim to further elucidate the content of Section~\ref{sec:explanation-decomposition} by presenting attribution maps for the decomposed explanations of both multiple features and multiple objects. Figure~\ref{fig:visualization-multiple-features-with-attribution} illustrates the generated explanations for an input characterized by multiple features, while Figure~\ref{fig:visualization-multiple-objects-with-attribution} depicts the explanations for an input containing several objects.

The algorithm for multiple path construction assumes that neurons in the higher layers have high-level features like the fox's eye. We generated multiple paths that include each hidden neuron extracted in Algorithm 1 from the top layer. For example, if a $path^{(N)} {=} \{h_1, h_2\}$ is generated at step $n{=}N$ in Algorithm 1, we assumed $path^{(N)} {=} \{h_1\}$ and proceeded to the next step to create a path. We also assumed $path^{(N)} {=} \{h_2\}$ and proceeded to the next step to create another path.

\section{Common Path Per Class}

\begin{figure}[!t]
    \centering
    \includegraphics[width=0.5\columnwidth]{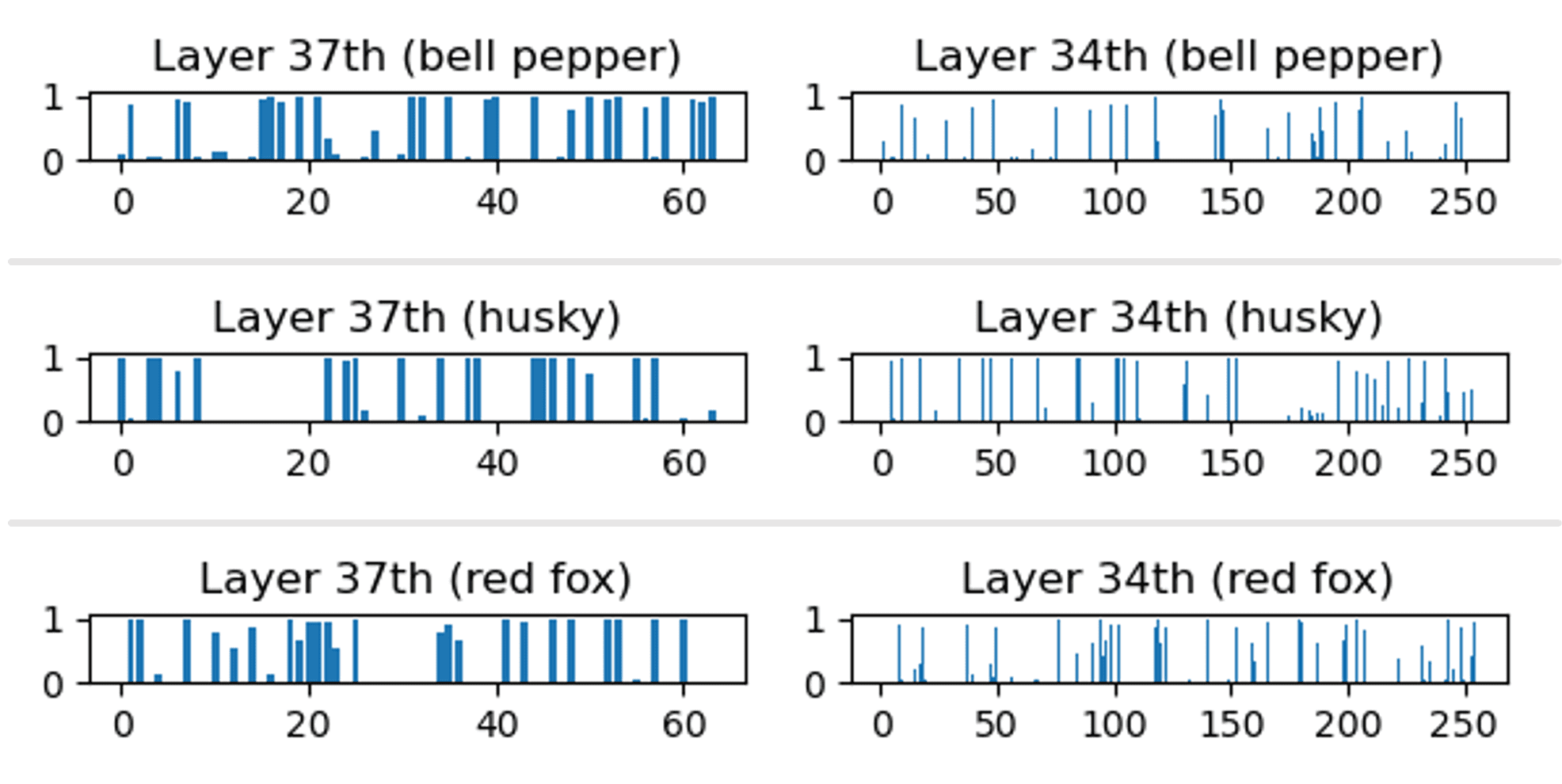}
    \vspace{-2mm}
    \caption{The frequency of hidden units for each layer included in paths per class when paths are respectively configured for all data on the curated subset of the ImageNet dataset. The x-axis represents the index of hidden units in that layer, and the y-axis represents the frequency ratio at which units of that index are selected. For better visualization, only the $37^{th}$ and $34^{th}$ layers are provided, and only three classes (\textit{bell pepper}, \textit{husky}, and \textit{red fox}) are provided.}
    \label{fig:commoncls}
\end{figure}

This section describes the consistent explainability of the proposed method by revealing the presence of frequently selected hidden units for each class, i.e., a common path for every class. As depicted in Figure~\ref{fig:commoncls}, the frequency with which hidden units are selected for path configuration is notably polarized (either extremely low or high) across all data points for each class in the dataset. This observation suggests that, for every layer and class, certain specific indices are chosen with high frequency, while others are rarely selected or not at all. Moreover, regardless of whether the classes are visually distinct (e.g., \textit{bell pepper} and \textit{husky}) or bear visual similarities (e.g., \textit{husky} and \textit{red fox}), the frequently selected hidden units differ significantly. In other words, a common path is uniquely associated with a specific class. Consequently, these findings underscore the capability of our pathwise explanation method to reveal the presence of a distinct path integral to the decision-making process for each class. This further bolsters the argument that our proposed method, which leverages only a subset of hidden units, offers a more consistent explanation compared to other methods that utilize all units.

\clearpage
\section{Further Comparison via a Different Architecture: ResNet} \label{sec:appendix_resnet}
Our pathwise method, in addition to the VGG architecture as discussed in the main script, can also be utilized to explain the decision-making process within another well-known architecture, ResNet. Unlike VGG, ResNet contains the residual connection, which operates based on the summation. However, when the path through all layers is modeled, it becomes apparent that the piecewise linear model for the path in ResNet is identical to that in VGG. In fact, by unfolding the formula as in Equation~\ref{eq:ReLU-NN-equation} and then extracting terms with $\phi$ for all hidden units as in Equation~\ref{eq:bpath}, this can be revealed.

\begin{figure*}[!t]
    \centering
    \includegraphics[width=0.8\textwidth]{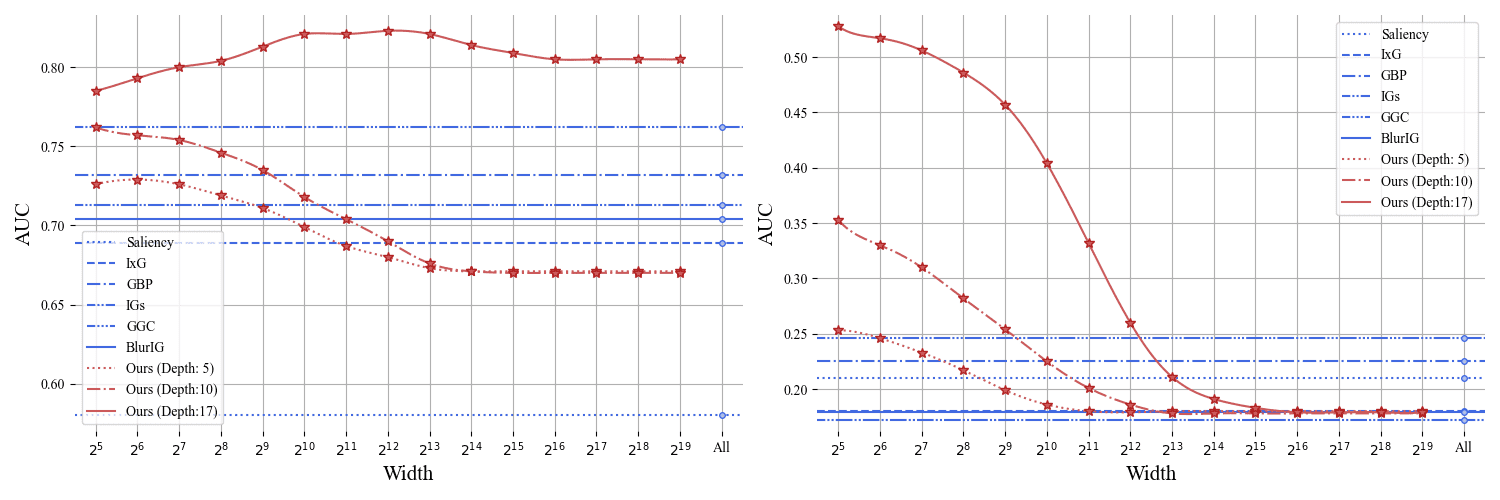}
    \vspace{-4mm}
    \caption{Different Architecture (ResNet) Experiment: Quantitative explanation via \textit{insertion} and \textit{deletion} depending on the maximum number of units per layer (\textit{width}) and the number of layers (\textit{depth}) for path configuration. In our method, a \textit{depth} of 17, which is the maximum, represents a complete path, while any lesser value indicates an incomplete path. }
    \label{fig:ins_perf_by_wd_res}
\end{figure*}

\begin{figure*}[!t]
    \centering
    \includegraphics[width=0.75\textwidth]{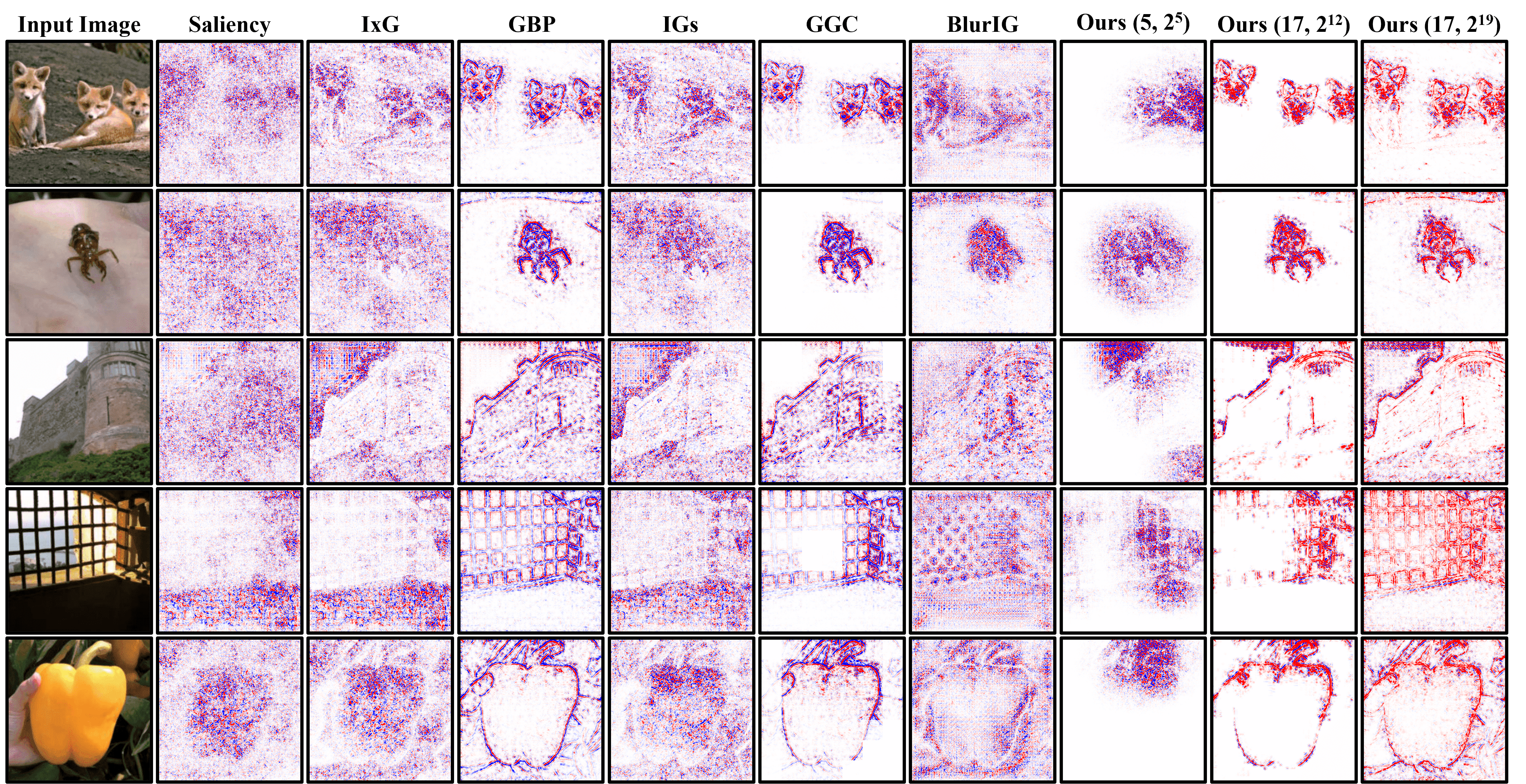}
    \vspace{-2mm}
    \caption{Different Architecture (ResNet) Experiment: Qualitative explanation depending on the maximum number of units per layer (\textit{width}) and the number of layers (\textit{depth}) for path configuration. The expression enclosed in parentheses in the proposed method is as follows: (\textit{depth}, \textit{width}). Positive and negative attributions are indicated in red and blue.}
    \label{fig:vis_resnet}
\end{figure*}

Table~\ref{tab:pixel insertion and deletion game resnet}, Figure~\ref{fig:ins_perf_by_wd_res}, and Figure~\ref{fig:vis_resnet} show the results of experiments with ResNet-18~\citep{he2016deep}, using the same settings as Table~\ref{tab:pixel insertion and deletion game}, Figure~\ref{fig:ins_perf_by_wd }, and Figure~\ref{fig:visualization special case} in the main script. Consequently, our method surpasses the others in the insertion metric and provides competitive performance in the deletion metric, even within the ResNet architecture. Furthermore, our study illustrates that the trends in path configuration, concerning changes in \textit{width} and \textit{depth}, align with those observed in VGG.

\begin{table}[t!]
    \centering
    \begin{tabular}{c|cccccc|c}
     & Saliency & IxG & IGs & GBP & GGC & BlurIG & Ours \\ \midrule
    \textit{Insertion}\,($\uparrow$) & 0.580 & 0.689 & 0.713 & 0.732 & 0.762 & 0.704 & \textbf{0.805} \\
    \textit{Deletion}\;($\downarrow$) & 0.210 & 0.180 & \textbf{0.172} & 0.225 & 0.246 & 0.181 & 0.179
    \end{tabular}
    \vspace{-2mm}
    \caption{Different Architecture (ResNet) Experiment: Area Under Curve (AUC) of \textit{Insertion} and \textit{Deletion}. $\uparrow$ indicates that the larger the value, the better the explanation, whereas $\downarrow$ indicates the opposite.}
    \label{tab:pixel insertion and deletion game resnet}
\end{table}

\clearpage

\begin{figure*}[!t]
    \centering
    \includegraphics[width=0.8\textwidth]{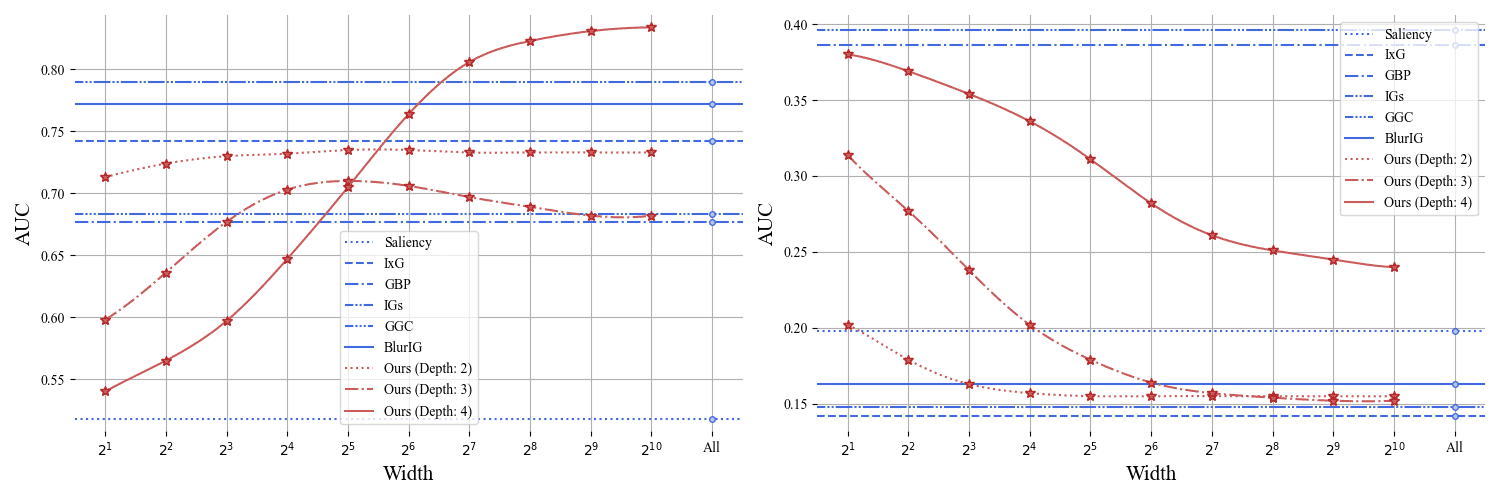}
    \vspace{-4mm}
    \caption{Low-Resolution Classification (CIFAR10) Experiment: Quantitative explanation via \textit{insertion} and \textit{deletion} depending on the maximum number of units per layer (\textit{width}) and the number of layers (\textit{depth}) for path configuration. In our method, a \textit{depth} of 4, which is the maximum, represents a complete path, while any lesser value indicates an incomplete path \vspace{-2mm}}
    \label{fig:wd_graph_cifar}
\end{figure*}

\begin{table}[t!]
    \centering
    \begin{tabular}{c|cccccc|c}
     & Saliency & IxG & IGs & GBP & GGC & BlurIG & Ours \\ \midrule
    \textit{Insertion}\,($\uparrow$) & 0.518 & 0.742 & 0.790 & 0.677 & 0.683 & 0.772 & \textbf{0.833} \\
    \textit{Deletion}\;($\downarrow$) & 0.198 & \textbf{0.142} & 0.148 & 0.386 & 0.396 & 0.163 & 0.239
    \end{tabular}
    \vspace{-2mm}
    \caption{Low-Resolution Classification (CIFAR10) Experiment: Area Under Curve (AUC) of \textit{Insertion} and \textit{Deletion}. $\uparrow$ indicates that the larger the value, the better the explanation, whereas $\downarrow$ indicates the opposite.\vspace{-2mm}}
    \label{tab:pidg_cifar}
\end{table}

\begin{figure*}[!t]
    \centering
    \includegraphics[width=0.75\textwidth]{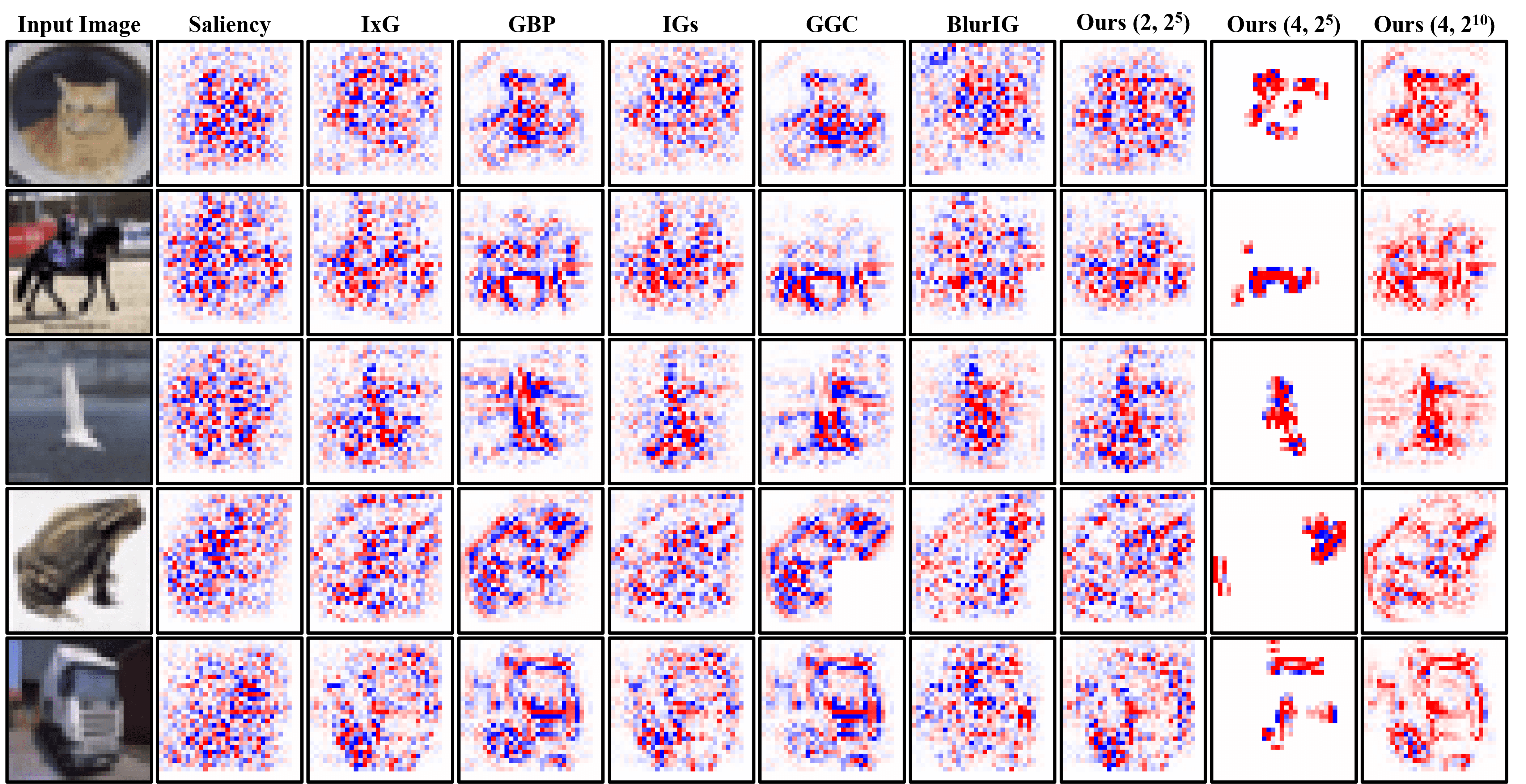}
    \vspace{-2mm}
    \caption{Low-Resolution Classification (CIFAR10) Experiment: Qualitative explanation depending on the maximum number of units per layer (\textit{width}) and the number of layers (\textit{depth}) for path configuration. The expression enclosed in parentheses in the proposed method is as follows: (\textit{depth}, \textit{width}). Positive and negative attributions are indicated in red and blue.}
    \label{fig:vis_cifar}
\end{figure*}

\section{Further Comparison in Low-Resolution Classification: CIFAR10} \label{further_comp}

This section provides additional quantitative comparisons performed on the CIFAR10 dataset~\citep{krizhevsky2009learning}. These comparisons were omitted from the main manuscript due to space constraints. CIFAR10 is a low-resolution dataset comprising 10 distinct classes. For our experiments on this dataset, we employed a toy model with three convolutional layers (each with a kernel size of 3) and two fully connected layers, offering an intuitive explanation suitable for its small resolution.

Even on CIFAR10, our method surpasses other approaches in terms of the \textit{insertion} metric, as evidenced in Table~\ref{tab:pidg_cifar}. Moreover, Figures~\ref{fig:wd_graph_cifar} and \ref{fig:vis_cifar} indicate that the proposed pathwise method delivers both quantitatively and qualitatively appropriate explanations for the CIFAR10 dataset. Specifically, as depicted in Figure~\ref{fig:wd_graph_cifar}, larger \textit{width} and \textit{depth} values can improve the upper bound of the \textit{insertion} metric, echoing the observations made in the main manuscript. However, our explanations for the \textit{deletion} metric show a slightly different trend when low-level layers are incorporated into the path due to a high \textit{depth} value. This deviation arises because the object-to-image size ratio in this dataset is substantial, given the image's very low resolution. Contrary to the \textit{insertion} metric, where pixels are inserted into a blank image factoring in the image's local area via low-level layers, in the \textit{deletion} metric, surrounding pixels can influence the model's decision when pixels are removed from the original image, considering the image's local area through low-level layers. Nonetheless, it's worth noting that our method still demonstrates competitive performance in the \textit{deletion} metric when adjusting the \textit{depth}. Figure~\ref{fig:vis_cifar} visually showcases our attribution maps, influenced by both \textit{depth} and \textit{width}. Due to the image's limited resolution, attributions cover the entire area when the \textit{depth} is minimal. Yet, it's evident that essential features remain prominently highlighted when the \textit{depth} is maximized, especially with appropriate \textit{width} adjustments.

\end{document}